\documentclass[acmtog]{acmart}

\usepackage{booktabs} 

\usepackage{microtype}
\usepackage{soul}
\usepackage{xspace}
\usepackage[ruled]{algorithm2e} 

\SetAlFnt{\small}
\SetAlCapFnt{\small}
\SetAlCapNameFnt{\small}
\SetAlCapHSkip{0pt}

\newcommand{\revision}[1]{\textcolor{black}{#1}}
\newcommand{\name}{StyleFlow\xspace}

\acmJournal{TOG}

\renewcommand\footnotetextcopyrightpermission[1]{}
\begin{document}


\title{StyleFlow:  
Attribute-conditioned Exploration of StyleGAN-Generated Images using 
Conditional Continuous Normalizing Flows}


\author{Rameen Abdal}
\affiliation{ \institution{KAUST} }
\email{rameen.abdal@kaust.edu.sa}

\author{Peihao Zhu}
\affiliation{ \institution{KAUST} }
\email{peihao.zhu@kaust.edu.sa}

\author{Niloy J. Mitra}
\affiliation{ \institution{UCL, Adobe Research} }
\email{nimitra@adobe.com}

\author{Peter Wonka}
\affiliation{ \institution{KAUST} }
\email{pwonka@gmail.com}


\fancyfoot{}

\begin{abstract}

High-quality, diverse, and photorealistic images can now be generated by unconditional GANs (e.g., StyleGAN). However, limited options exist to control the generation process using (semantic) attributes, while still preserving the quality of the output. Further, due to the entangled nature of the GAN latent space, performing edits along one attribute can easily result in unwanted changes along other attributes. In this paper, in the context of \textit{conditional exploration} of entangled latent spaces, we investigate the two sub-problems of attribute-conditioned sampling and attribute-controlled editing. We present StyleFlow as a simple, effective, and robust solution to both the sub-problems by formulating  conditional exploration as an instance of conditional continuous normalizing flows in the GAN latent space conditioned by attribute features. We evaluate our method using the face and the car latent space of StyleGAN, and demonstrate fine-grained disentangled edits along various attributes \revision{ on both real photographs and StyleGAN generated images}. For example, for faces we vary camera pose, illumination variation, expression, facial hair, gender, and age. Finally, via extensive qualitative and quantitative comparisons, we demonstrate the superiority of StyleFlow to other concurrent works. \\ Project Page : \textbf{\url{https://rameenabdal.github.io/StyleFlow}}
\\ Video : \textbf{\url{https://youtu.be/LRAUJUn3EqQw}}

\end{abstract}

%
%

\ccsdesc[300]{Neural Rendering~Face Editing}
\ccsdesc[300]{Flows~Continuous Normalizing Flows}
\ccsdesc[300]{Generative Modeling~ GANs}

%
%

\keywords{Generative Adversarial Networks, Image Editing}

\begin{teaserfigure}
\centering
\includegraphics[width=\linewidth]
			{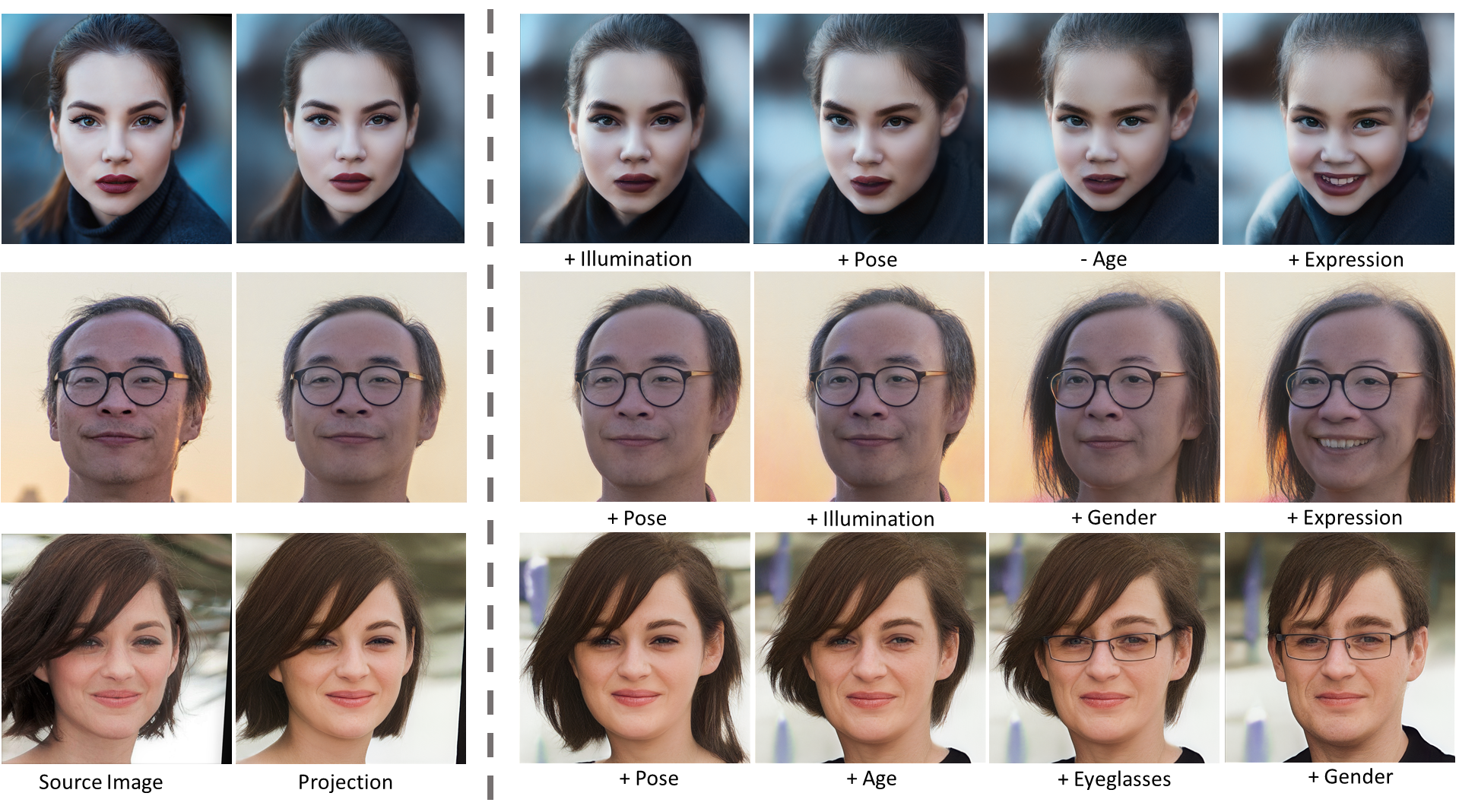}
\caption{We present \textit{StyleFlow} to enable attribute-conditioned semantic edits on projected real images and  StyleGAN generated images. For each of these examples, the user sequentially changes (camera) pose, illumination, expression, eyeglasses, gender, and age of a real image. Please judge, where applicable,  the extent of identity preservation of the respective person under the applied edits. In this figure, all the source images are real images.  }
\label{fig:teaser}
\end{teaserfigure}

\maketitle

\thispagestyle{empty}


\section{Introduction}

A longstanding goal of Computer Graphics has been to generate high-quality realistic images that can be controlled using user-specified attributes. One broad philosophy has been to create detailed 3D models, decorate them with custom materials and texture properties, and render them using realistic camera and illumination models. Such an approach, once realized, provides users with significant control over a range of attributes including object properties, camera position, and  illumination. However, it is still hard to achieve photorealism over a range of attribute specifications.

Alternately, in a recent development, generative adversarial networks~(GANs) have opened up an entirely different image generation paradigm. Tremendous interest in this area has resulted in rapid improvements both in speed and accuracy of the generated results. For example, StyleGAN~\cite{Karras_2019}, one of the most celebrated GAN frameworks, can produce high resolution images with unmatched photorealism. However, there exist limited options for the user to control the generation process with adjustable attribute specifications. For example, starting from a real face photograph, how can we edit the image to change the (camera) pose, illumination, or the person's expression?

One successful line of research that supports some of the above-mentioned edit controls is \textit{conditional generation} using GANs. In such a workflow, attributes have to be specified directly at (GAN) training time. While the resultant conditional GANs~\cite{choi2019stargan,Xiao_2018_ECCV, park2019SPADE} provide a level of (semantic) attribute control, the resultant image qualities can be blurry and fail to reach the quality produced by uncontrolled GANs like StyleGAN. Further, other attributes, which were not specified at training time, can change across the generated images, and hence result in loss of (object) identity due to edits.

We take a different approach. We treat the attribute-based editing problem as a \textit{conditional exploration} problem in an unsupervised GAN,  rather than conditional generation requiring attribute-based retraining. We explore the problem using StyleGAN, as one of the leading uncontrolled GAN setups, and treat attribute-based user manipulations as finding corresponding {\em non-linear paths} in the StyleGAN's latent space. Specifically, we explore two problems: (i)~\textit{attribute-conditioned sampling}, where the goal is to sample a diverse set of images all meeting user-specified attribute(s); and (ii)~\textit{attribute-controlled editing}, where the user wants to edit, possibly sequentially, a particular image with target attribute specifications. 
The paths inferred by StyleFlow is conditioned on the input image, and hence can be adapted to the uniqueness of individual faces.

Technically, we solve the problem by proposing a novel normalizing flow based technique to conditionally sample from the GAN latent space.
    First, assuming access to an attribute evaluation function (e.g., an attribute classification network), we generate sample pairs linking StyleGAN latent variables with attributes of the corresponding images.  In our implementation, we consider a range of attributes including camera, illumination, expression, gender, and age for human faces; and camera, type, and color for cars. We then enable adaptive latent space vector manipulation by casting the conditional sampling problem in terms of conditional normalizing flows using the attributes for conditioning. Note that, unlike in conditional GANs, this does \textit{not} require attribute information during GAN training. This results in a simple yet robust attribute-based image editing framework.

\begin{figure}[b!]
        \centering
        \includegraphics[width=\linewidth]{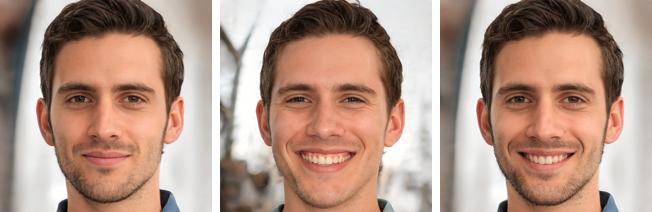}
        \caption{Given an input StyleGAN image~(left), vector arithmetic in the latent space~\cite{RadfordMC15} (here computed by Image2StyleGAN~\cite{abdal2019image2stylegan}) can produce expression edits~(middle) but at the cost of changing identity of the face due to the entangled latent space. In contrast, our proposed StyleFlow~(right), by extracting non-linear paths in the latent space, enables expression edits while retaining identity.  }
        \label{fig:motfig}
    \end{figure}

\begin{figure*}[t!]
        \centering
        \includegraphics[width=\linewidth]{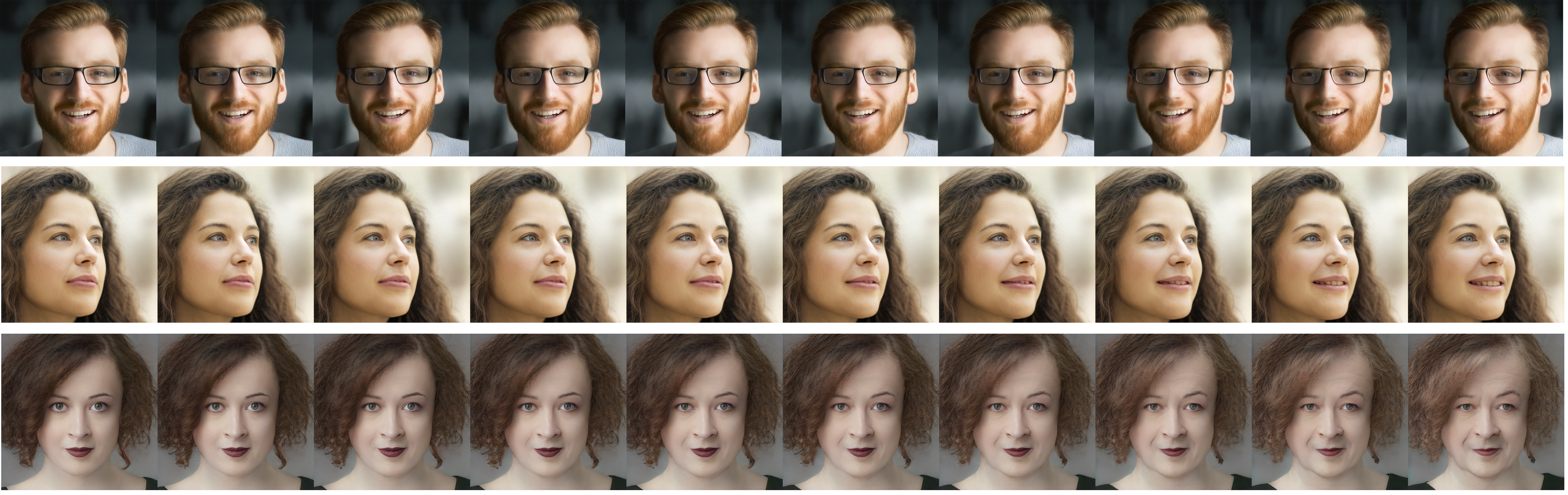}
        \caption{Disentangled edits performed by our StyleFlow framework on real images. From top to bottom, changing camera view, expression, and age, respectively. Please note the non local changes. For example, in the second row, the eyes and cheeks respond under increasing smile.}
        \label{fig:motfig2}
    \end{figure*}

We evaluate our method mainly in the context of human faces (both generated as well as real images projected into the relevant latent space) and present a range of high-quality \textit{identity-preserving} edits at an unmatched quality (contrast against Figure~\ref{fig:motfig} produced using latent space arithmetic). As a further proof of robustness, we demonstrate sequential edits, see Figure~\ref{fig:teaser} and Figure~\ref{fig:seqedit}, to the images without forcing the latent vectors out of the distribution, as guaranteed by our formulation of the problem with normalizing flows. We compare our results with concurrently proposed manipulation techniques, and demonstrate superior identity preservation, both quantitatively and qualitatively.
In summary, we enable conditional exploration of latent spaces of unconditional GANs using conditional normalizing flows based on (semantic) attributes.

Project code and user interface (see supplementary material) will be released for research use.

\section{Related Work}
\label{sec:related-work}

\paragraph{Generative Adversarial Network Architecture}
Generative Adversarial Networks (GANs) have been introduced by Goodfellow et al.~\shortcite{goodfellow2014generative} and sparked a huge amount of follow up work.
One direction of research has been the improvement of the GAN architecture, loss functions, and training dynamics. As current state of the art we consider a sequence of architecture versions by Karras and his co-authors, ProgressiveGAN~\cite{karras2017progressive}, StyleGAN~\cite{Karras_2019}, and StyleGAN2~\cite{karras2019analyzing}. These architectures are especially strong in human face synthesis.
Another strong system is BigGAN~\cite{brock2018large} that produces excellent results on ImageNet~\cite{imagenet_cvpr09}. We build our work on StyleGAN2 as it is easier to work with. A detailed review of the history of GAN architectures, or a discussion on loss functions, etc. is beyond the scope of the paper. 
We also note that in addition to GANs, there are other generative models, such as Variational Autoencoders (VAEs)~\cite{kingma2013autoencoding} or pixelCNN~\cite{oord2016conditional}. While these methods have some advantages, GANs currently produce the highest quality output for the applications we consider by a large margin.

\paragraph{Conditional GANs}
A significant invention for image manipulations are conditional GANs (CGANs).
To add conditional information as input, CGANs  \cite{Mirza2014ConditionalGA} learn a mapping $G : {x,z} \rightarrow y$ from
an observed input $x$ and a randomly sampled vector $z$ to an output image $y$.
One important class of CGANs use images as conditioning information,
such as pix2pix~\cite{Isola_2017}, BicycleGAN~\cite{zhu2017toward}, pix2pixHD~\cite{wang2018pix2pixHD}, SPADE~\cite{park2019SPADE}, MaskGAN~\cite{lee2019maskgan} and SEAN~\cite{zhu2019sean}. \revision{Other notable works~\cite{nie2020semisupervised,aliaks2020order, Zakharov_2019} produce high quality image animations and manipulations.}
CGANs can be trained even with unpaired training data using cycle-consistency loss~\cite{zhu2017unpaired, yi2017dualgan, kim2017learning}.
They can be used as a building block for image editing, for example, by using a generator G to translate a line drawing or semantic label map to a realistic-looking output image. CGANs have given rise to many application specific modifications and refinements.


\paragraph{Applications of Conditional GANs}
CGANs are an excellent tool for semantic image manipulations. In the context of faces, StarGAN~\cite{Choi_2018,choi2019stargan} proposes a GAN architecture that considers face attributes such as hair color, gender, and age.
Another great idea is to use GANs conditioned on sketches and color information to fill in regions in a face image. This strategy was used by FaceShop~\cite{10.1145/3197517.3201393} and SC-FEGAN~\cite{Jo_2019_ICCV}. One interesting aspect of these papers is that they use masks to restrict the generation of content to a predefined region. Therefore, some of the components of these systems borrow from the inpainting literature. We just mention DeepFillv2~\cite{yu2018free} as a representative state-of-the-art technique. 
An early paper in computer graphics showed how conditioning on sketches can produce good results for terrain modeling~\cite{guerin2017interactive}.
Specialized image manipulation techniques, such as makeup transfer PSGAN~\cite{jiang2019psgan} or hair editing~\cite{MichiGAN} are also very useful, as faces are an important class of images.
A very challenging style transfer technique is the transformation of input photographs to obtain caricatures~\cite{Cao_2019}. This problem is quite difficult, because the input and output are geometrically deformed.
Overall, these methods are not directly comparable to our work because the input and problem statements are slightly different.

\paragraph{Image Editing by Manipulating Latent Codes}
A competing approach to conditional GANs is the idea to manipulate latent codes of a pretrained GAN. Radford et al.~\shortcite{RadfordMC15} observed that interesting semantic editing operations can be achieved by first computing a difference vectors between two latent codes (e.g. a latent code for a person with beard and a latent code for a person without beard) and then adding this difference vector to latent codes of other people (e.g. to obtain an editing operation that adds a beard).
Our technique falls into this category of methods that do not design a separate architecture, but manipulate latent codes of a pretrained GAN. This approach became very popular recently and all noteworthy competing papers are only published on arXiv at the point of submission. We would like to note that these papers were developed independently of our work. Nevertheless, we believe it will be useful for the reader to judge our work in competition with these recent papers \cite{nitzan2020disentangling,tewari2020stylerig,harkonen2020ganspace}, because they provide better results than other work.
A great idea is proposed by the StyleRig~\cite{tewari2020stylerig} paper where they want to transfer face rigging information from an existing model as a method to control face manipulations in the StyleGAN latent space. While the detailed control of the face ultimately did not work, they have very nice results for the transfer of overall pose (rotation) and illumination.
Based on earlier work, some authors worked on the hypothesis that the StyleGAN latent space is actually linear and they propose linear manipulations in that space. Two noteworthy efforts are InterFaceGAN~\cite{shen2019interpreting} and GANSpace~\cite{harkonen2020ganspace}. The former tries to find the latent space vectors that correspond to meaningful edits. The latter takes a data driven approach and uses PCA to learn the most important directions. Upon analyzing these directions the authors discover that the directions often correspond to meaningful semantic edits. Our results confirm that the assumption of a linear latent space is a useful simplification that produces good results. However, we are still able to produce significantly better \revision{disentangled} results with a non-linear model of the latent space \revision{conditioned on the input image}.

\paragraph{Embedding Images into the GAN Latent Space}
We would also like to mention techniques that try to embed images into the latent space of a GAN. Generally, there are three main techniques. The first technique is to build an encoder network that maps an image into the latent space. The second technique is to use an optimization algorithm to iteratively improve a latent code so that it produces the output image~\cite{abdal2019image2stylegan, karras2019analyzing, Abdal_2020_CVPR}. There is the idea to combine the two techniques and first use the encoder network to obtain an approximate embedding and then refine it with an optimization algorithm~\cite{zhu2016generative, zhu2020indomain}. Finally, other methods~\cite{zhu2019disentangled} use VAEs to create inverse mappings. 
We will use the optimization based technique of Karras et al.~\cite{karras2019analyzing}.
In addition, embedding can itself be used for GAN-supported image modifications. We will compare to one recent approach in our work~\cite{abdal2019image2stylegan}.


\paragraph{Neural Rendering}
Neural rendering refers to the idea to generate images from a scene description using a neural network. We refer the reader to a recent state of the art report~\cite{Tewari2020NeuralSTAR} that summarizes recent techniques.
Current methods tackle specific sub-problems like novel view synthesis~\cite{hedman2018deep, thies2020image, sitzmann2019deepvoxels}, relighting under novel lighting conditions~\cite{xu2018deep, guo2019relightables},  animating faces~\cite{kim2018deep, thies2019deferred,Fried}, and animating bodies~\cite{aberman2019deep, shysheya2019textured, martin2018lookingood} in novel poses. While these techniques share some similar goals in terms of user interaction, the overall problem setting is sufficiently different from our work so that it is hard to compare to these works directly.









\begin{figure*}[ht]
        \centering
        \includegraphics[width=\linewidth]{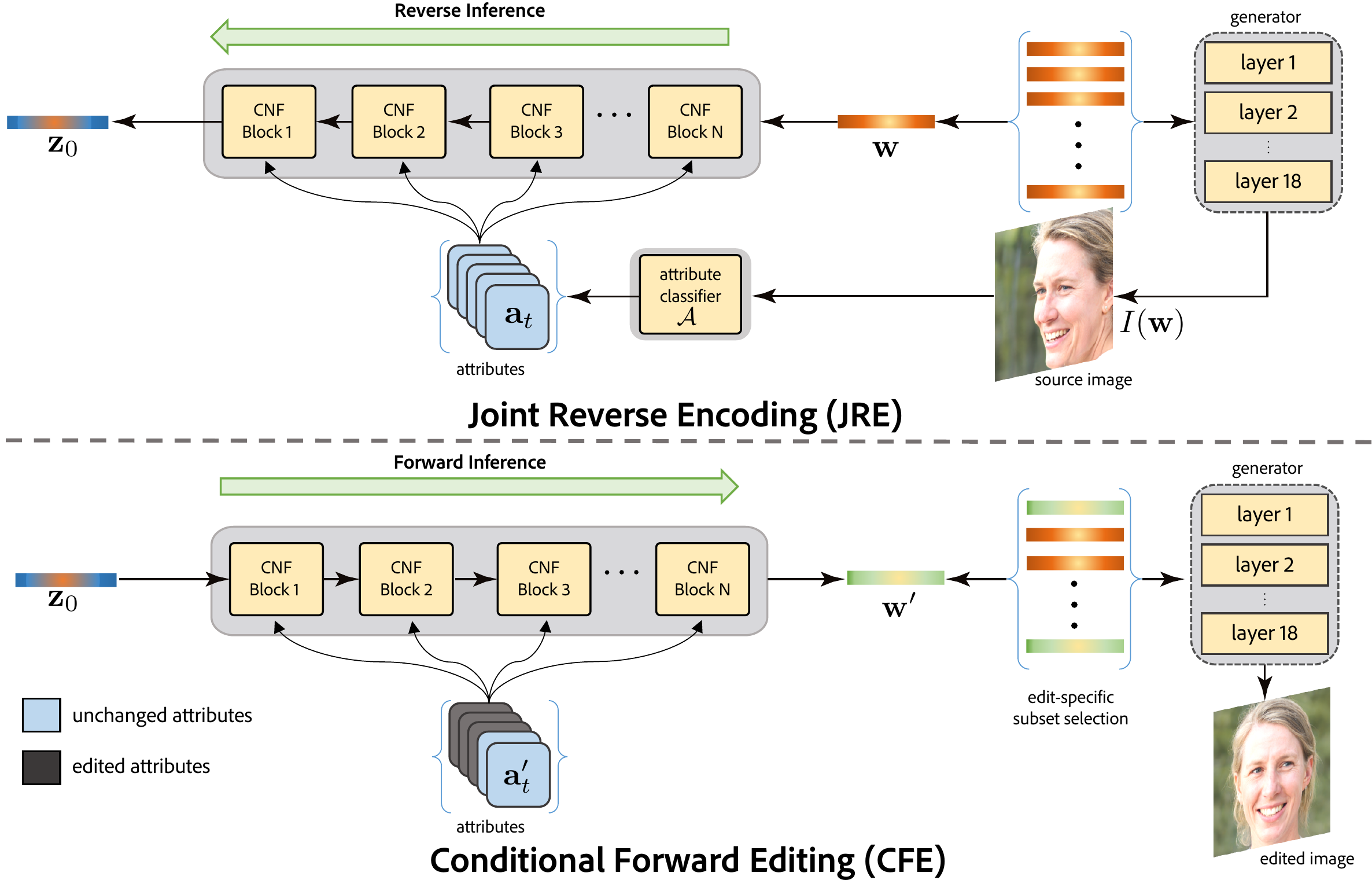}
        \caption{Attribute-conditioned editing using StyleFlow. Starting from a source image, we support attribute-conditioned editing by using a reverse inference followed by a forward inference though a sequence of CNF blocks (see Figure~\ref{fig:illustration2}). Here, $\mathbf{z}$ denotes the variable of the prior distribution and $\mathbf{w}$ denotes the intermediate weight vector of the StyleGAN. \revision{Also note that the reverse and the forward inferences are solved by an ODE solver by evaluating CNF functions over the time variable.} Please refer to the text for details. }
        \label{fig:illustration1}
    \end{figure*}

\section{Overview}

We support two tasks: first, {\em attribute-conditioned sampling}, wherein we want to sample high-quality realistic images with target attributes; and, second, {\em attribute-controlled editing}, wherein we want to edit given images such that the edited images have the target attributes, while best preserving the identity of the source images. 

For generating realistic images, we use StyleGAN. In our implementation, we support sampling from both StyleGAN~\cite{Karras_2019} and StyleGAN2~\cite{karras2019analyzing}. We recall that StyleGAN maps latent space samples $\mathbf{z}_s \in \mathbb{R}^{512}$ to intermediate vectors $\mathbf{w} \in \mathbb{R}^{512}$ in $W$ space,  by learning a non-linear mapping $f: \mathbf{z}_s \rightarrow \mathbf{w}$, such that the $\mathbf{w}$-s can then be decoded to 
images $I(\mathbf{w}) = I(f(\mathbf{z}_s)) \in \mathbb{R}^{3 \times 1024 \times 1024}$. In the uncontrolled setup, $\mathbf{z}_s$ is sampled from a multi-dimensional normal distribution. 
\revision{ The $\mathbf{w}$ vector is used to control the normalization at 18 different locations of the StyleGAN2 generator network. The idea of the extended latent space $W+$ is to not use the same vector $\mathbf{w}$ eighteen times, but use different vectors. Hence, a vector $\mathbf{w} \in W+$ has dimensions $18 \times 512$. We will use both of these latent spaces in our paper. For training the StyleFlow network, we use $W$. For restricting edits and editing real images, we use $W+$.}

In order to measure attributes of any image, we assume access to a class-specific attribute function $\mathcal{A}$, typically realized as a classifier network, that returns a vector of attributes $\mathbf{a}:= \mathcal{A}(I)$ for any given image $I$ belonging to the class under consideration. The attributes are represented as an $l$-dimensional vector  (e.g., $l=17$ for human faces in our tests). 

Solving the first task amounts to sampling $\mathbf{z}$ from a multi-dimensional normal distribution and using a learned mapping function of the form  $\Phi(\mathbf{z},\mathbf{a})$, where $\mathbf{a}$ denotes the target attributes, to produce suitable intermediate weights. These weights, when decoded via StyleGAN, produce attribute-conditioned image samples of the form $I(\Phi(\mathbf{z},\mathbf{a}))$ matching the target attribute. 

Specifically, using a zero-mean multi-dimensional normal distribution with identity as variance we can conditionally sample as, 
\begin{equation}
    \mathbf{z} \sim N(\mathbf{0}, \mathcal{I}) \;\; \text{and} \;\; \mathbf{w} =  \Phi(\mathbf{z},\mathbf{a}) 
\end{equation} 
and in the process satisfy  $\mathcal{A}(I(\Phi(\mathbf{z},\mathbf{a})) = \mathbf{a}$. 
In Section~\ref{sec:method}, we describe how to train and use a neural network to model such a function $  \Phi(\mathbf{z},\mathbf{a})$ using forward inference on a conditional continuous normalizing flow~(CNF). In other words, the normalizing flow maps the samples from an n-dimensional prior distribution, in this case a normal distribution, to a latent distribution conditioned on the target attributes.

Solving the second task is more complex. Given an image $I_0$, we first project it to the StyleGAN space to obtain $\mathbf{w}_0$ such that $I(\mathbf{w}_0) \approx I_0$ using \cite{abdal2019image2stylegan, karras2019analyzing}. 
Recall that our goal is to edit the current image attributes $\mathbf{a}_0 = \mathcal{A}(I(\mathbf{w}_0))$ to user specified attributes $\mathbf{a}_t$, whereby the user has indicated changes to one or multiple of the original attributes, while best preserving the original image identity. 
We then recover latent variables $\mathbf{z}_0$ that lead to intermediate weights $\mathbf{w}_0$ using an inverse lookup
$\mathbf{z}_0 = \Psi(\mathbf{w}_0, \mathbf{a}_0)$. We realize the inverse map using a reverse inference of the CNF network described earlier, i.e., $\Psi(\mathbf{w}_0, \mathbf{a}_0) := 
\Phi^{-1}(\mathbf{w}_0, \mathbf{a}_0)$. Finally, we perform a forward inference, using the same CNF, to get the edited image $I_t$ that preserves the identity of the source image as, 
\begin{equation}
I_t = \Phi( \mathbf{z}_0, \mathbf{a}_t ) = 
\Phi( \Phi^{-1}(\mathbf{w}_0, \mathbf{a}_0), \mathbf{a}_t )
= 
\Phi( \Phi^{-1}(\mathbf{w}_0,  \mathcal{A}(I(\mathbf{w}_0)) ), \mathbf{a}_t ). 
\end{equation}
We first summarize normalizing flows in Section~\ref{sec:CNF}, and then, in Section~\ref{sec:method}, we describe how the invertible CNF can be used to compute the exact likelihood of the samples from the latent distribution of a generative model.

\section{Normalizing Flows}
\label{sec:CNF}

A normalizing flow, often realized as a sequence of invertible transformations, allows to map an unknown distribution to a known distribution (e.g., normal or uniform distribution). This inverse mapping, from an initial density to a final density and vice versa, can be simply seen as a chain of recursive change of variables.

\subsection{Discrete Normalizing Flows}
Let $\phi: \mathbb{R}^{d} \rightarrow \mathbb{R}^{d}$ be a  bijective map such that there exists an invertible map $g$ with $g := \phi^{-1}$. Let the transformation of the random variable be from $\mathbf{z} \sim p_z(\mathbf{z})$ to $\mathbf{w}$ such that $\mathbf{w} = \phi(\mathbf{z})$. By the change of variable rule, the output probability density of variable $\mathbf{w}$ can be obtained as,
\begin{equation}p_{w}\left(\mathbf{w}\right)=p_{z}(\mathbf{z})\left|\operatorname{det} \frac{\partial \phi}{\partial \mathbf{z}}\right|^{-1}\end{equation}
where $\phi^{-1}(\mathbf{w}) = \mathbf{z}$ or $g(\mathbf{w}) = \mathbf{z}$. The same rule applies for a successive transformation of the variable $\mathbf{z}$. Specifically,  the transformation be represented by $\mathbf{w} = \phi_{K}\left(\phi_{K-1}\left(\ldots \phi_{1}(\mathbf{z}_{0})\right)\right)$, i.e., 
$\mathbf{z}_0 \rightarrow \ldots \mathbf{z}_{K-1} \rightarrow \mathbf{z}_K = \mathbf{w}$, 
and since $\phi^{-1}$ exists the inverse mapping is expressed as  $\mathbf{z}_{0} = \phi_{1}^{-1}\left(\phi_{2}^{-1}\left(\ldots \phi_{K}^{-1}(\mathbf{w})\right)\right)$. Therefore, applying the change of variable rule, we obtain the modified output log probability density, 
\begin{equation}\log p_{w}\left(\mathbf{w}\right)=\log p_z\left(\mathbf{z_{0}}\right)-\sum_{n=1}^{K} \log \operatorname{det}\left|\frac{\partial \phi_{n}}{\partial \mathbf{z}_{n}}\right|,\end{equation}
where $\mathbf{z}_{n+1} = \phi_{n}(\mathbf{z}_{n})$ and $\mathbf{z}_{K} = \mathbf{w}$.

In the special case of planar flows, the function $\phi$ can be modeled by a neural network~\cite{rezende2015variational} where the flow takes the form, 
\begin{equation}\mathbf{z}_{n+1}=\mathbf{z}_{n}+ \mathbf{u}_n h \left(\mathbf{w}_n^{\top} \mathbf{z}_{n}+b\right),\end{equation}
where $\mathbf{u}_n \in \mathbb{R}^d, \mathbf{w}_n \in \mathbb{R}^d, b \in \mathbb{R}$ are the learnable parameters, $h()$ is a smooth element-wise non-linear activation with derivative $h'()$. 
The probability density obtained by sampling $p_z(\mathbf{z_{0})}$ and applying a sequence of planar transforms to produce variable $\mathbf{w} = \mathbf{z}_K$  takes the form,  
\begin{equation}\log p_{w}\left(\mathbf{w}\right)=\log p_z\left(\mathbf{z_{0}}\right)-\sum_{n=1}^{K} \log \left|1+\mathbf{u}_n^{\top} \xi(\mathbf{z}_{n-1}) \right|.\end{equation} 
where $\xi(\mathbf{z}) = h' (\mathbf{w}^\top \mathbf{z} + b) \mathbf{w}$.

\begin{figure}[t!]
        \centering
        \includegraphics[width=\columnwidth]{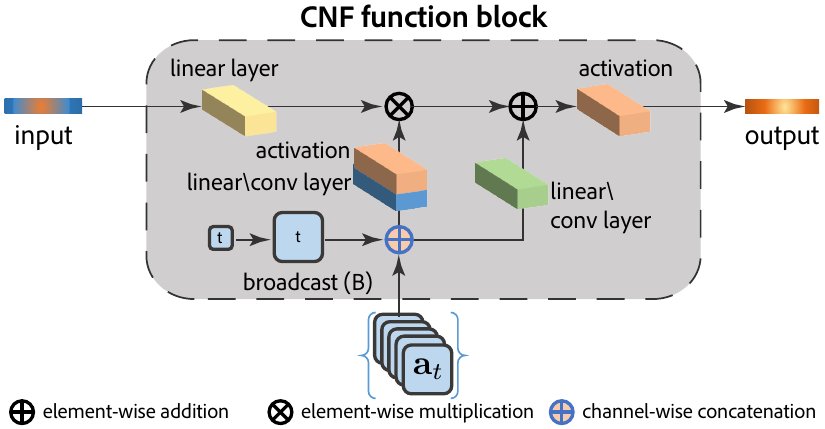}
        \caption{Conditional continuous normalizing flow~(CNF) function block realized as a neural network block. Note that the learned function, conditioned on attribute vector $\mathbf{a}_t$, can be used for both forward and backward inference. }
        \label{fig:illustration2}
    \end{figure}

\subsection{Continuous Normalizing Flows~(CNF)} 
The normalizing flows can be generalized into a continuous formulation~\cite{grathwohl2018ffjord,chen2018neural} using neural ODE~\cite{chen2018neural} which adopts adjoint sensitivity method~\cite{pontryagin2018mathematical} to compute the gradients with respect to the parameters in an ODE black box solver. In continuous Normalizing flows,  differential equations are expressed in the form: $\frac{d \mathbf{z}}{d t}=\phi(\mathbf{z}(t), t; \theta)$ where $z$ is the variable of a given distribution, $t$ is the time variable, and $\theta$ are the parameters of an arbitrary neural network. Specifically, the differential equation takes the form, 
\begin{equation}z(t_{1})=z\left(t_{0}\right)+\int_{t_{0}}^{t_{1}} \phi(z(t), t;\theta) d t, \nonumber  \end{equation} 
Finally, the change in the log density can be expressed as, 
\begin{equation}\log p\left(\mathbf{z}\left(t_{1}\right)\right)=\log p\left(\mathbf{z}\left(t_{0}\right)\right)-\int_{t_{0}}^{t_{1}} \operatorname{Tr}\left(\frac{\partial \phi}{\partial \mathbf{z}(t)}\right) d t. \end{equation}

\revision{We decided against using DNF networks as is that it is difficult to ensure a reversible mapping. Another related problem is the expressiveness and versatility of such networks due to fixed number of invertible functions to choose from. Finally, the Jacobian determinant computation is costly, so a workaround is to constrain the network architecture which is also undesirable~\protect\cite{grathwohl2018ffjord}. }

\revision{In our StyleFlow framework, we use CNFs for our formulation. 
In contrast, the main benefit of the continuous formulation is that it is invertible by definition and the determinant is replaced by a matrix trace, which is considerably easier to compute. Hence, it allows to choose from a wide variety of architectures. Additionally, FFJORD~\protect\cite{grathwohl2018ffjord} also claims that CNFs can potentially learn a less entangled internal representation compared to DNFs (see e.g. Figure 2 in FFJORD).
 We would also like to point out that in this work, we interpret time as a "virtual" concept related to how CNFs are evaluated. Instead of evaluating the results by sequentially passing through the network layers, as in DNF, in CNFs the ODEs are used to evaluate the function through time. Hence, for both our conditional sampling and editing tasks, we desire to condition based on the target attributes and use CNFs to continuously evolve the image samples. The  CNFs are expected to retarget the probability densities as described next.}

\section{Method}
\label{sec:method}

We consider the latent vectors $\mathbf{w} \in \mathbb{R}^{512}$ sampled from the $W$ space of the StyleGAN1/2. The prior distribution is represented by ${p}_z(\mathbf{z})$, where $\mathbf{z} \in \mathbb{R}^{512}$ . Our aim is to model a conditional mapping between the two domains. Moreover, it is imperative to be able to learn a semantic mapping between the domains so that editing applications are realizable. We explain our method in the following subsections.

\subsection{Dataset Preparation}
A general work flow for the preparation of the dataset is as follows: We first sample 10k samples from the Gaussian $\mathbb{Z}$ space of the StyleGAN1/2~\cite{karras2019analyzing,Karras_2019}. Then we infer the corresponding $\mathbf{w}$ codes in the disentangled $W$ space of the models. We use the vectors $\mathbf{w}$ of $W$ space truncated by a factor of $0.7$  as suggested by StyleGAN. \revision{Otherwise, there will be outlier faces of low quality}.
We generate corresponding images $I(\mathbf{w})$ via StyleGAN1/2 generator and hence create a mapping between the $W$ space and the image space $I$. To have conditional control over the image features, we use a face classifier network $\mathcal{A}$  to map the images $I$ to the attribute $A_{t}$ domain. We use this dataset for the final training of the flow network using triplets $w\in W$, $i \in I$ and $a_{t} \in A_t$. \revision{ Note that the attributes $a_t$ do not depend of the variable $t$.}         
To prepare the $A_t$ domain of the training dataset, we use a state-of-the-art Microsoft Face API~\cite{API}, which we found to be robust for the face attribute classification. The API provides a diverse set of attributes given a face image. The main attributes that we focus in our work are gender, pitch, yaw, eyeglasses, age, facial hair, expression, and baldness. For the lighting attribute, we use the predictions from the DPR  model~\cite{DPR} that outputs a 9-dimensional vector per image measuring the first 9 spherical harmonics of the lighting. Thus, for faces, we have $a_t \in \mathbb{R}^{17}$.

\begin{figure*}[ht]
        \centering
        \includegraphics[width=\linewidth]{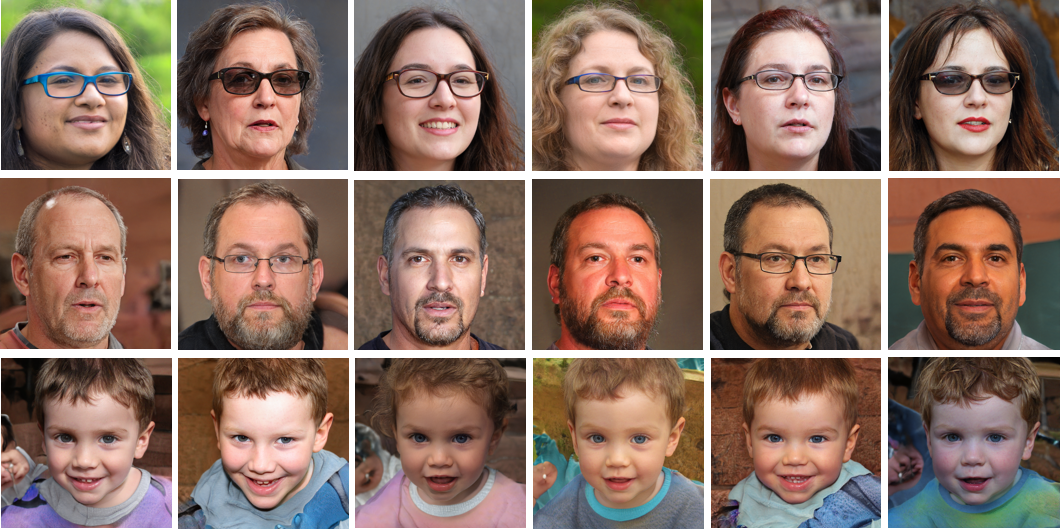}
        \caption{
        Attribute-conditioned sampling using StyleFlow by resampling $z_0$ given the attributes. Here we show sampling results for attribute specifications of \textit{females with glasses in a target pose}~(top); \textit{50-year old males with facial hair}~(middle); and \textit{smiling 5-year old children in a target pose}~(bottom). }
        \label{fig:sampling}
    \end{figure*}

\subsection{Attribute-translation Model}
We use a series of the gate-bias modulation~\cite{grathwohl2018ffjord, pointflow} (called "ConcatSquash") networks to model the function $\phi$ of the conditional continuous normalizing flows. \revision{ We make this design choice as the gate part can learn the per-dimension scaling factor given an input latent code and the bias part of the network can learn to translate the code towards a particular edit direction which is suitable for our formulation of adaptive identity aware edits. This model builds on top of FFJORD \protect\cite{grathwohl2018ffjord} as a general framework where the attributes can be 2D or 3D tensors like an image for an Image2Image translation task.} Figure~\ref{fig:illustration2} shows the function block used in the CNF block. To include the condition information into the network, we transform the time variable $t$ with a broadcast operation $B$ to match the spatial dimensions of the attribute space. Then, we apply channel-wise concatenation to the resultant variable with the attribute variable $a_t$, and finally the new variable ${a^{+}_t}$ is fed to the network as a conditional attribute variable. Note that at inference time, we use linear interpolation in the attribute domain to smoothly translate between the two edits to get the final image. \revision{For stable training, we used 4 stacked CNF functions (i.e., gate-bias functions) and 2 Moving Batch norm functions~\protect\cite{pointflow} (one before and one after the CNF functions) where each function outputs a vector of the same dimension (i.e. 512). We observed that using only 2-3 CNF function block models overfit on the data. The matrix trace is computed by 10 evaluations of Hutchinson’s trace estimator.}
Depending on the properties of the extended ${a^{+}_t}$ tensor, we can use the convolutional or linear neural network to transform the tensors to make them the same shape as the input. \revision{We make use of linear layers in this work, but we expect extensions to this work where the flows can be conditioned on images, segmentation maps etc.} Then, we perform gate-bias modulation on the input tensor. Note here we use sigmoid non-linearity before the gate tensor operation~\cite{grathwohl2018ffjord}. The final output tensor is passed through a $Tanh$ non-linearity before passing to the next stage of the normalizing flow.  

An important insight of our work is that flow networks trained on one attribute at a time learn entangled vector fields, and hence resultant edits can produce unwanted changes along other attributes. Instead, we propose to use joint attribute learning for training the flow network. We concatenate all the attributes to a single tensor before feeding it to the network. In Section~\ref{sec:results}, we show that the joint attribute training produces an improvement in the editing quality with a more disentangled representation. We hypothesize that the training on single condition tends to over-fit the model on the training data. Further, in absence of measures along other attribute axis, the conditional CNF remains oblivious of variations along those other attribute directions. Therefore, the flow changes multiple features at a time. Joint attribute training tends to learn stable conditional vector fields for each attribute.

\subsection{Training Dynamics}

The goal during the training is to maximize the likelihood of the data $\mathbf{w}$ given a set of attributes $a_t$ . So, the objective can be written as \revision{$max_\theta \prod_{w,a_t} p(w|a_t,\theta)$.} Here, we assume the standard Gaussian prior with $z$ as the variable. Also, let  $\mathcal{N}$ represent the Gaussian probability density function. Algorithm~\ref{alg:one} shows the training algorithm~\cite{grathwohl2018ffjord, chen2018neural} of the proposed joint conditional continuous normalizing flows. \revision{Other details are Epochs: 10; Batch size: 5; Training speed: 1.07 - 2.5 iter/sec depending on the number of CNF functions (see Table~\ref{tab:abb}); GPU: Nvidia Titan XP; Parameters: 1128449; Final Log-Likelihood: -4327872; Inference time: 0.61 sec. For faster implementation and practical purposes, we also train a model with 6 stacked CNF functions which improves the inference time to 0.21 sec at the cost of slight decrease in the quality of disentanglement. We use the adjoint method to compute the gradients and solve the ODE using `dopri5' ODE solver~\protect\cite{grathwohl2018ffjord}. The tolerances are set to default $1\times10^{-5}$. } 
 We use the Adam optimizer with an initial learning rate of $1\times10^{-3}$. Other parameters $(\beta_1, \beta_2)$ of the Adam optimizer are set to default values.

\begin{algorithm}[h!]
\SetAlgoNoLine
\KwIn{Paired latent-attribute data ($\{w,a_{t}\}$); Neural network $\phi$; Integration times $t_{0}$ and $t_{1}$; ODE solver with adjoint sensitivity method; Number of training steps $N_{t}$; Optimizer $F'$; Learning rate $\eta$. }
\textbf{Initialization:}\\
${\left[\begin{array}{c}\mathbf{z}\left(t_{1}\right) \\ \log p(\mathbf{w}|a_{t})-\log p \left(\mathbf{z}\left(t_{1}\right)\right)\end{array}\right]=\left[\begin{array}{l}\mathbf{w} \\ 0\end{array}\right]}$
; ${a_{t}^{+}}$ = $B(t)$ || $a_{t}$, \\
where $B$ expands the variable $t$ such that spatial dimension of $a_{t}$ is equal to $t$ and || is the concatenation operation.

        \For{ i = [1 : $N_{t}$]
    }{$\left[\begin{array}{c}\mathbf{z_{0}}\\ \mathbf{\Delta_{logp}} \end{array}\right]=\int_{t_{1}}^{t_{0}}\left[\begin{array}{c}\phi(\mathbf{z}(t), {a_{t}^{+}} ; \theta) \\ -\operatorname{Tr}\left(\frac{\partial \phi}{\partial \mathbf{z}(t)}\right)\end{array}\right] d t$ \\ 
    $\mathcal{L} = \log{\mathcal{N}(z_{0}; 0, I)} - \Delta_{logp}$
    
    $ \theta \leftarrow \theta - \eta F'(\nabla_{\theta}\mathcal{L},\theta)$\;
        }
\caption{Flow training Algorithm}
\label{alg:one}
\end{algorithm}

\begin{figure*}[t!]
        \centering
        \includegraphics[width=\linewidth]{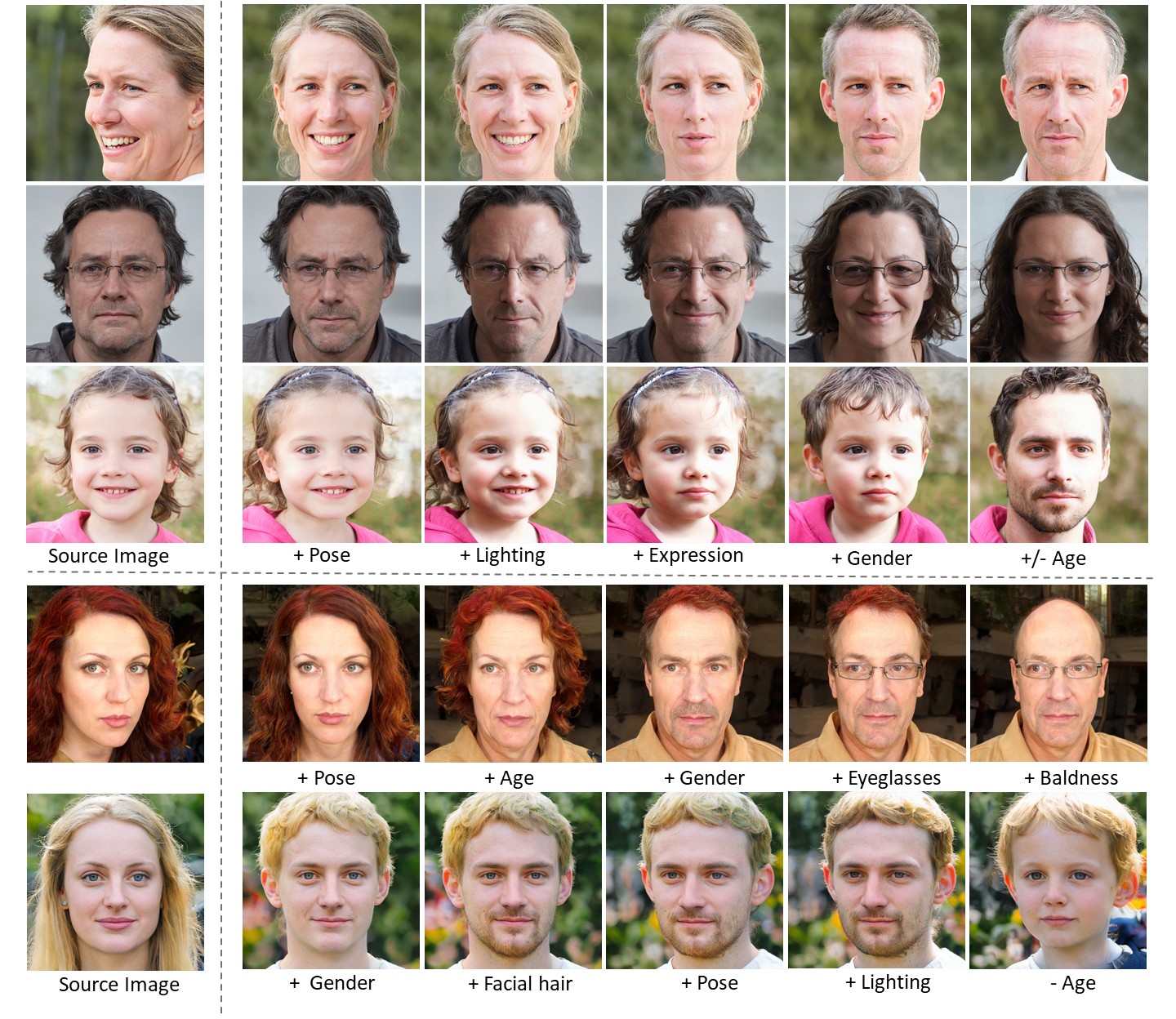}
        \caption{Sequential edits using StyleFlow on synthetically generated faces with `+'/`-' denoting corresponding attribute was increased/decreased. Please notice the quality of preservation of attributes that are not being edited, demonstrating the  disentanglement of the various attributes.  }
        \label{fig:seqedit}
    \end{figure*}

\section{Attribute-conditioned Sampling and Editing}
A natural benefit of the training formulation of the framework is the sampling. In particular, the mapping learnt between the two domains is used to produce a vector $z$ given a $w$ vector and vice versa. Moreover, we can manipulate the vectors in the respective domains and the changes translate to the other domain semantically from the editing point of view. Please refer to the supplementary video for interaction sessions. 

\subsection{Conditional Sampling}
Once the network is trained, we are able to conditionally sample the $w \in W$ with the Gaussian prior modelled by the continuous normalizing flows. Formally, we set the attribute variable $a_t$ to a desired set of values, and then sample multiple $z \sim p(z)$. These vectors are then passed through the (trained) conditional CNF network. The learned vector field translates the vectors to produce the latent vectors $w$, which are then finally fed to the StyleGAN1/2 generator. In Section~\ref{sec:results}, we show the results of sampling given a set of attributes. We notice that the quality of the samples is very high as well as unedited attributes remain largely fixed. The conditional sampling is an important result of our work and validates the fact that the network is able to learn the underlying semantic representations,  which is further used to perform semantic edits to the images.


\subsection{Semantic Editing}
Here we show the procedure to semantically edit the images using the proposed framework. Note here the vector manipulations are adaptive and are obtained by solving a vector field by an ODE solver. Unlike the previous methods~\cite{abdal2019image2stylegan,harkonen2020ganspace, shen2019interpreting} the semantic edits performed on the latent vectors $w$ forces the resultant vector to remain in the distribution of $W$ space ($p(w)$). This enables us to do stable sequential edits which, to the best of our knowledge, are difficult to obtain with the previous methods. We show the results of the edits in Section~\ref{sec:results}. In the following, we will discuss the procedure and components of the editing framework. 

\subsubsection{Joint Reverse Encoding (JRE)} The first step of the semantic editing operation in the StyleFlow framework is the Joint Reverse Encoding. Here, we jointly encode the variables $w$ and $a_t$. Specifically, given a $w \in W$, we infer the source image $i \in I$. Note that we can also start with a real image and use the projection methods~\cite{abdal2019image2stylegan, karras2019analyzing,zhu2020indomain},  to infer the corresponding $w$. Such procedures may render the vectors outside the original $p(w)$ distribution and hence makes the editing a challenging task. \revision{Later, we show that StyleFlow is a general framework that also works on real images.} We pass the image $I$ through the face classifier API~\cite{API} and the lighting prediction DPR network~\cite{DPR} to infer the attributes. Then, we use reverse inference given a set $w$ and $a_t$ to infer the corresponding $z_0$.

\subsubsection{Conditional Forward Editing (CFE)} The second step is the Conditional Forward Editing, where we fix the $z_0$ \revision{(this vector encodes a perfect projection of the given image)} and to translate the semantic manipulation to the image $I$, we change the set of desired conditions, e.g.,  we change the age attribute from 20 yo to 60 yo. Then, with the given vector $z_0$ and the new set of (target) attributes $a'_t$, we do a forward inference using the flow model. Finally, we process the resulting vector $w'$ to produce an editing vector.

\subsubsection{Edit Specific Subset Selection} This is the third step of the StyleFlow editing framework. Studying the structure of the StyleGAN1/2, we choose to apply the given vector $w'$ at the different indices of the $W+$\cite{Karras_2019, abdal2019image2stylegan,Abdal_2020_CVPR} ($\mathbb{R}^{18\times512}$) space depending on the nature of the edit, e.g.,  we would expect the lighting change to be in the later layers of the StyleGAN where mostly the color/style information is present. Empirically, we found the following indices of the edits to work best: Light ($7 - 11$), Expression ($4 - 5$), Yaw ($0 - 3$), Pitch ($0 - 3$), Age ($4 - 7$), Gender ($0 - 7$), Remove Glasses ($0 - 2$), Add Glasses ($0 - 5$), Baldness ($0 - 5$) and Facial hair ($5 - 7$ and $10$). The final step is the inference of the image from the modified latents. \revision{We refer the framework without the Edit Specific Subset Selection Block as V1 and the final framework is referred to as V2.} In Section~\ref{sec:results}, we show the importance of this module in improving the editing quality. 

\revision{We have two ways to edit: in the faster and approximate version, we do not reproject every time an edit is performed; and in the slower and accurate implementation, all the vectors (18 w-s) are reprojected in one pass after an edit. As the w code is manipulated every time by the w-s based on the edit-specific subset selection, some of these subsets overlap with others in a sequential edit and may make the edit unstable. In the fast option, occasionally there can be sudden jumps in the output image. Encoding these vectors back to space ensures that the flow network is aware of the changes made to the W/W+ space (identity-aware) of the StyleGAN1/2 and hence make the edits stable. Note that since the previous and concurrent methods are not identity aware, this problem also adds to the reason for failed sequential edits. So, in our work } the vectors of resultant W+ space are re-mapped to a new set of $z_0$s using JRE followed by CFE and edit specific subset selection to perform a subsequent edit.

\begin{figure*}[t!]
        \centering
        \includegraphics[width=\linewidth]{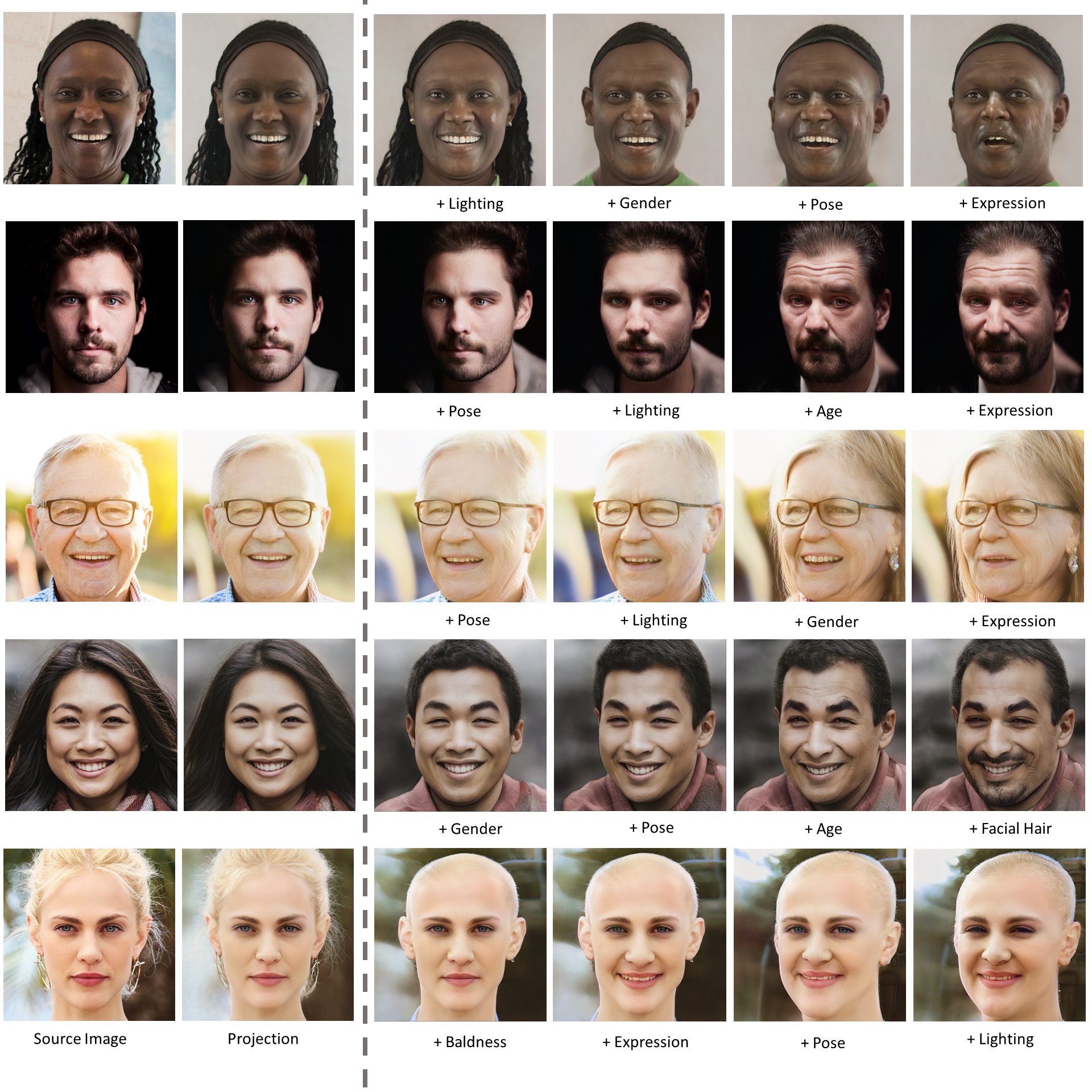}
        \caption{
       \revision{ Real image sequential edits using our StyleFlow framework. Note that we show different permutations of the edits to demonstrate high quality of a particular edit that can appear anywhere in the sequence.}
        }
        \label{fig:real_0}
    \end{figure*}

\begin{figure*}[t]
        \centering
        \includegraphics[width=\linewidth]{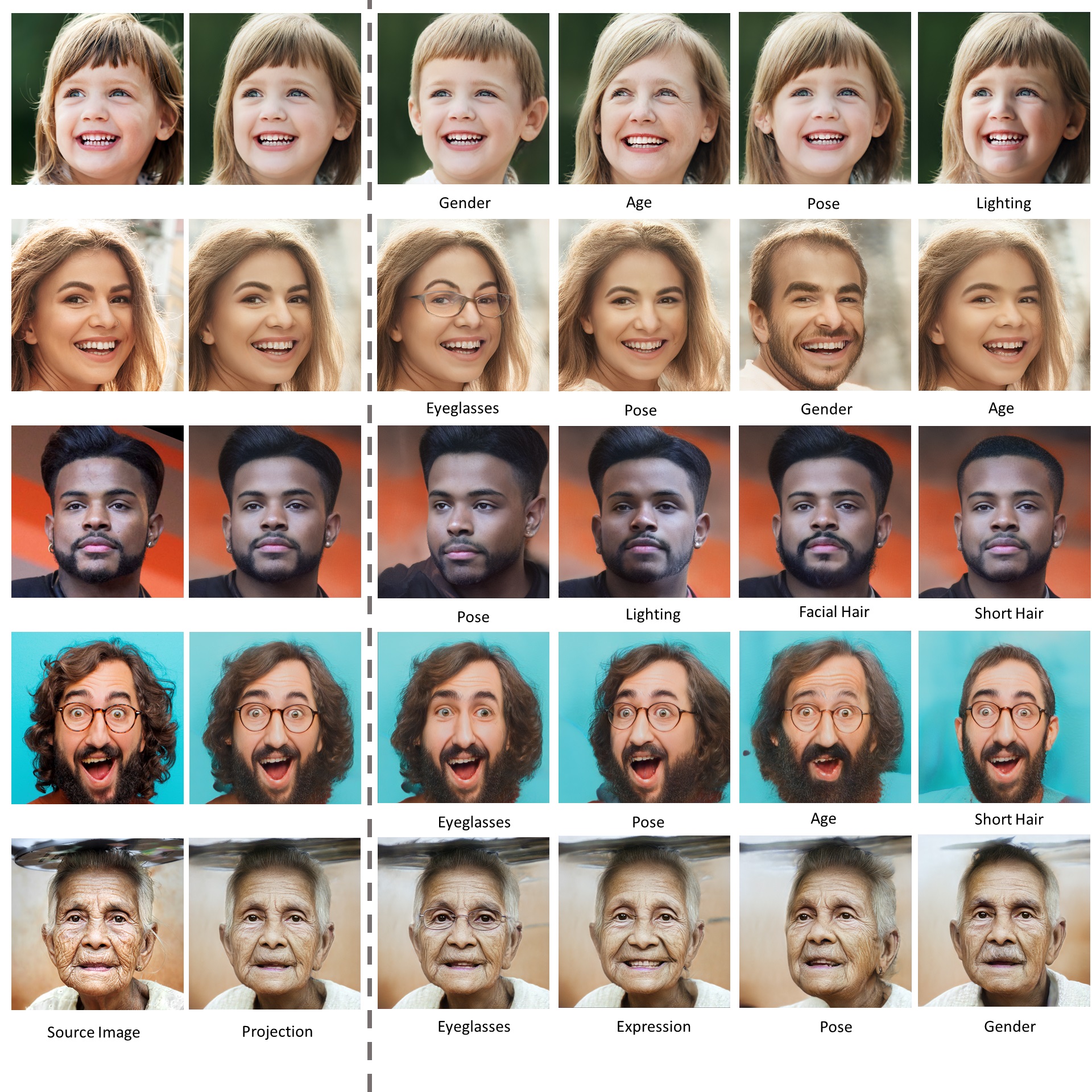}
        \caption{
   \revision{Real image non-sequential edits using our StyleFlow framework. Note that the method is able to handle extreme pose (first and second rows), asymmetrical expressions (fourth row) and age diversity (first and last rows) well compared to the concurrent methods. }
        }
        \label{fig:real2}
    \end{figure*}
    
\section{Results}
\label{sec:results}
In this section, we discuss the results produced by our StyleFlow framework and compare them, both quantitatively and qualitatively, with other methods.

\subsection{Datasets}

We evaluate our results on two datasets - FFHQ~\cite{Karras_2019} and LSUN-Car~\cite{yu15lsun}.   Flickr-Faces-HQ (FFHQ) is a $1024\times1024$ resolution high-quality image dataset of human faces consisting of 70,000 images, which are diverse in terms of ethnicity, age, and accessories. LSUN-Car is a $512\times384$ resolution image dataset of cars consisting of 16,185 images, which are  diverse in terms of car pose, color, and types. We use StyleGAN pretrained on these datasets to evaluate our results.

\subsection{Evaluation metrics}

We evaluate the results of our framework and competing methods using three different metrics namely FID, face identity, and edit consistency scores. 

\subsubsection{FID score:} To measure the diversity and quality of the output samples, we use the FID score between the test images and the edited images. We evaluate the results with 1k generated samples from the StyleGAN2 framework.

\subsubsection{Face identity score:} 
To measure the quality of the edit and quantify the identity preservation of the edits, we evaluate the edited images using a face identity score. 
We take a state-of-the-art classifier model for face recognition~\cite{Fr} to output embeddings of the images. Given a pair of images, before and after edits, we calculate the Euclidean distance and the cosine similarity between the embeddings. Note that we use a different classifier  from the attribute estimator used in training our StyleFlow.  We choose three major edits for this purpose: light, pose, and expression.

\subsubsection{Edit consistency score:}
To measure the consistency of the applied edit across the images, we evaluate over different edit permutations. For example, in a sequential editing setup, if the pose edit is applied, it should not be affected by where in the sequence it is applied. In principle, in the above case, different permutations of edits should lead to the same attributes when classified with an attribute classifier. 
Say,  $ep$ refers to an expression edit followed by a pose edit, while $pl$ refers to a pose edit followed by a lighting edit. We expect the pose attribute to be the same when evaluated on the final image. We measure this using the score 
$
| \mathcal{A}_p ( E_p(E_e(I)) - 
\mathcal{A}_p ( E_l(E_p(I)) |,
$
where 
$E_x$ denotes conditional edit along attribute specification $x$ and 
$\mathcal{A}_p$ denotes pose attribute vector regressed by the attribute classifier.

\subsection{Compared Methods}
\revision{We compare to a simple version of vector arithmetic as demonstrated in Image2StyleGAN. Additionally, we compare to three concurrent works: InterfaceGAN, GANSpace, and StyleRig.  
InterfaceGAN and Image2StyleGAN were retrained using StyleGAN2. GANSpace and StyleFlow are naturally implemented in StyleGAN2. However, StyleRig uses StyleGAN1.
}

\noindent {\bf (i) Image2StyleGAN:} For Image2StyleGAN, we embedded paired images of expression (IMPA-FACES3D~\cite{impa}), lighting pairs from DPR, and the rotation pairs using StyleFlow outputs. The lighting part of both Image2StyleGAN and InterfaceGAN is trained for  right-to-left illumination change.

\noindent {\bf (ii) InterfaceGAN~\cite{shen2019interpreting}:} We retrained InterfaceGAN on the same data as StyleFlow. The images were segregated based on the attributes to create the binary data. For the magnitude of edits, note that it is difficult for competing methods to produce the same magnitude of changes when a given vector is applied for different faces. For example, in Figure 1 supplementary materials, top row, the target rotation is a complete flip, for InterfaceGAN, the learned vector is translated till it matches the extreme pose. Hence, instead of evaluating the results by considering the attributes as binary, we use three continuous metrics, as previously described. 

\noindent {\bf (iii) GANSpace~\cite{harkonen2020ganspace}:} We used the code provided by the authors and use the version using layer subsets. \revision{Note that to match the edits for the generated images, we used the sigmas -15 for the expression (index 46), -3 for the pose (index 1) and 10 for the LR lighting direction (index 12) from the official GANSpace open source implementation. For real images, we use the following sigmas: Expression -15, Pose 2 and LR lighting direction 7 to match the edits.}

\noindent {\bf (iv) StyleRig~\cite{tewari2020stylerig}:} The comparison results used in the paper were prepared by the authors of StyleRig. \\
We would like to reiterate that the last three competing methods were only available on arXiv at the time of submission and were independently developed.

\subsection{Qualitative comparison of edits}

\revision{We show sequential edits on real images projected to StyleGAN2 by re-implementing the Image2StyleGAN W+ projection algorithm in Figures~\ref{fig:teaser} and \ref{fig:real_0} . In addition to the sequential edits, we show non-sequential edits in Figure~\ref{fig:real2}. Note that we demonstrate results on images with diverse pose, lighting, expressions, and age attributes. For example, in row 4 in Figure~\ref{fig:real2}, even though the expression is asymmetrical, \name handles the edits well. 
}

\begin{figure}[ht]
        \centering
        \includegraphics[width=\columnwidth]{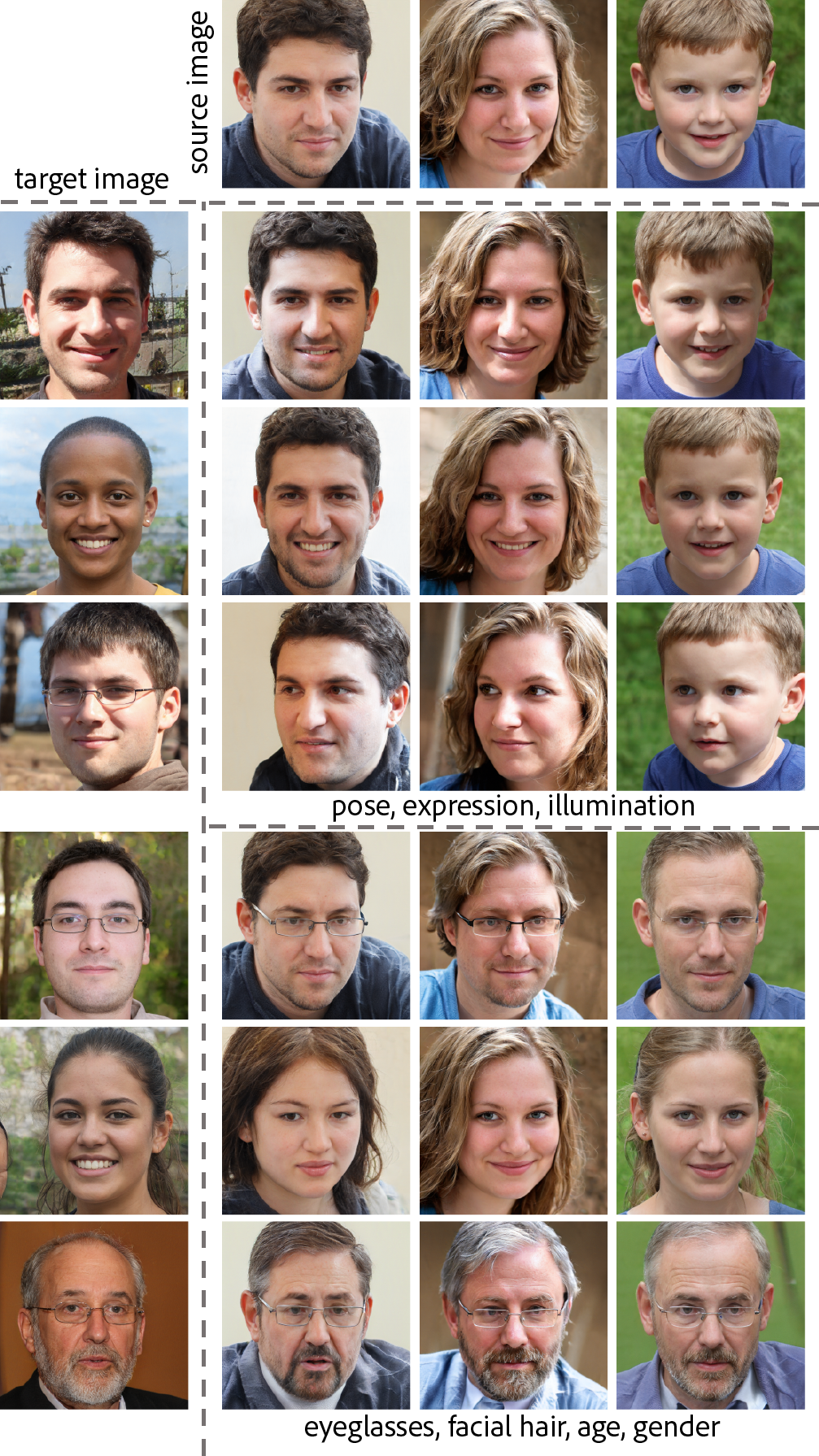}
        \caption{Attribute-conditioned edits where source images are edited using multiple attributes from the target images. Upper set uses pose, expression, and illumination from the respective target images; bottom set uses eyeglass, facial hair, age, and gender from the respective target images. }
        \label{fig:transfer}
    \end{figure}

 Figure~\ref{fig:seqedit} shows the results of the sequential edits on generated images using our framework. Here, we consider the sequential edits of Pose $\rightarrow$ Lighting $\rightarrow$ Expression $\rightarrow$ Gender $\rightarrow$ Age. \revision{ In order to show that different permutations of the edits can be performed without affecting the performance,  Figure~\ref{fig:seqedit} and Figure~\ref{fig:real_0} show the results of a random sequence of edits performed to a source image. Here we consider multiple edits of gender, facial hair, pose, lighting, age, expression, eyeglasses, and baldness. Note the quality of the edits. Also, notice that the order of the edits does not affect the quality of the images.} We can handle extreme pose changes while smoothly transferring the edits as the attributes change. Global features like background, cloths, and skin tone are largely preserved. Moreover, we also show high quality results for attribute transfer in Figure~\ref{fig:transfer}. 


Other approaches that directly manipulate the latent space, e.g., adding offset vectors, are not able to achieve the same quality because vector manipulations often move the final latent into a region outside the distribution~\cite{karras2019analyzing}. 
\revision{\name deviates from these methods in two ways. Firstly, attribute-guided edits amount to non-linear curves in the StyleGAN latent space. 
Secondly, the above curves (or even their linear approximations, see Figure~\ref{fig:interpol}) are conditioned on the current identity. Note that this is in contrast to edit vectors being independent of current identity as in GANSpace and InterfaceGAN (see Figures~\ref{fig:realx} and \ref{fig:real} for comparison). }

\revision{We also compare with the StyleRig method. StyleRig has not been designed to work with real images while \name can be applied to projected real images (see Figure~\ref{fig:teaser}, Figure~\ref{fig:real},  Figure~\ref{fig:realx} and Figure~\ref{fig:real2}). StyleRig also only supports a subset of edits. Additionally, while the sequential edits are important in practice and is one of the important contributions of this work, StyleRig does not work well in this setting. Figures~\ref{fig:realx} and \ref{fig:real} compare with results kindly produced by the StyleRig authors on our test scenario (4 representative images taken from real image dataset).} We show that the quality of the edits by our method on attribute transfer are of similar or higher visual quality. 

\subsection{Quantitative comparison of edits}
\label{sec:comparison}

First, we would like to analyze how much the edits depend on the initial latent vector. For an edit $w \rightarrow w'$ we can compute the difference vector between the final latent vector $w'$ and the initial latent vector $w$. The linear models Image2StyleGAN and InterfaceGAN make the assumption that the difference vectors are independent of the starting latent vector $w$ if the same edit is applied. We perform the following test to evaluate how much we deviate from this assumption.
Given many edits of the same type (e.g. changing a neutral expression to smiling by translating the attribute from 0 to 1), we compute their difference vectors $w'- w$. Then, given a set of pairs of these vectors, we compare the magnitude and the angles between the vectors. \revision{By sampling multiple such edits, the mean of the magnitudes (norm) of these difference vectors is computed to be $12.5$ indicating the adaptive nature of the edits}. The angles between the vectors are observed to vary up to ${36}^{\circ}$. This shows that the edits depend on the initial latent $w$ allowing the resultant vector to follow the original posterior distribution. Hence, we condition the edits wrt the source models to produce higher quality edits.

\begin{figure*}[t]
        \centering
        \includegraphics[width=\linewidth]{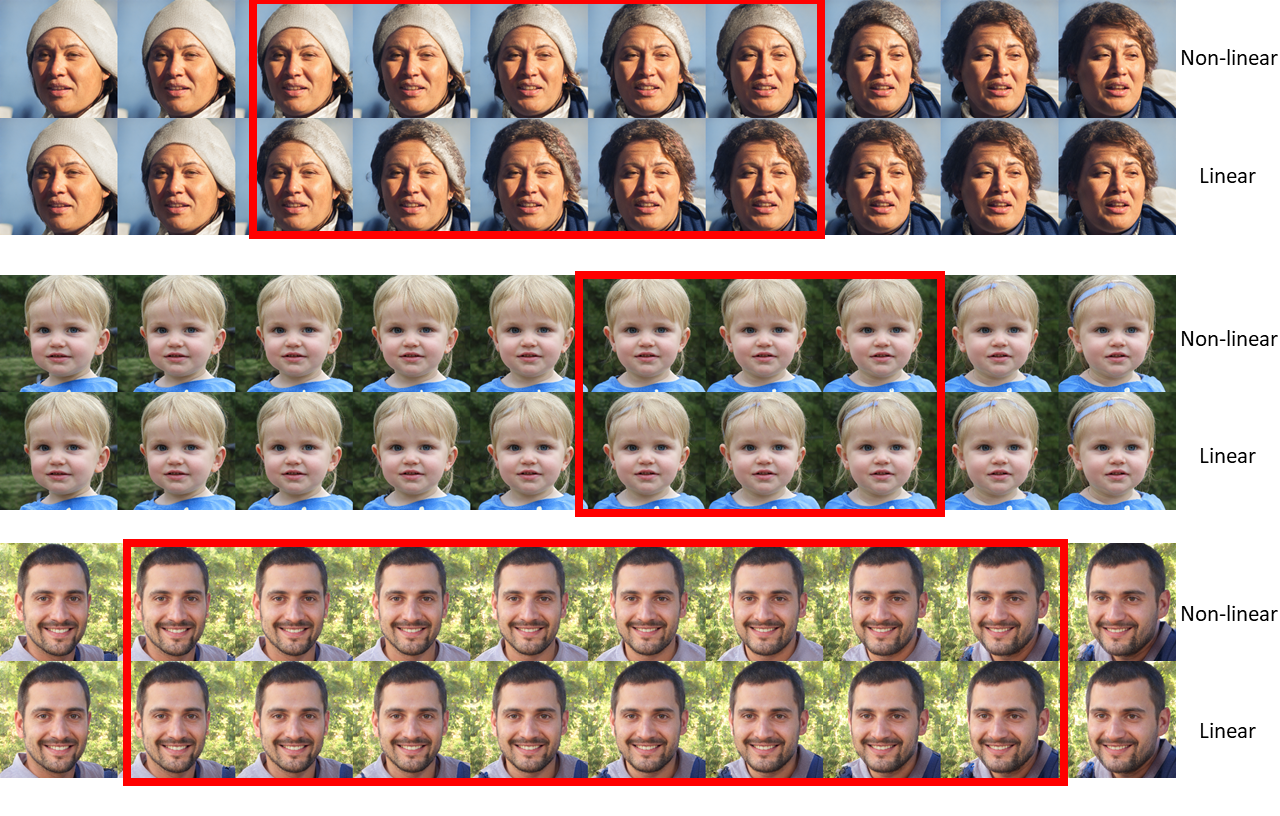}
        \caption{
    \revision{ Exploration of the non-linear vs linear edit paths produced by interpolating the StyleFlow variables $a_{t}$ vs $w$. Note here top row in every image is a non-linear path and the bottom row represents a linear path. Red box shows the region in which they differ significantly.  }  
        }
        \label{fig:interpol}
    \end{figure*}

\begin{figure}[t!]
        \centering
        \includegraphics[width=\linewidth]{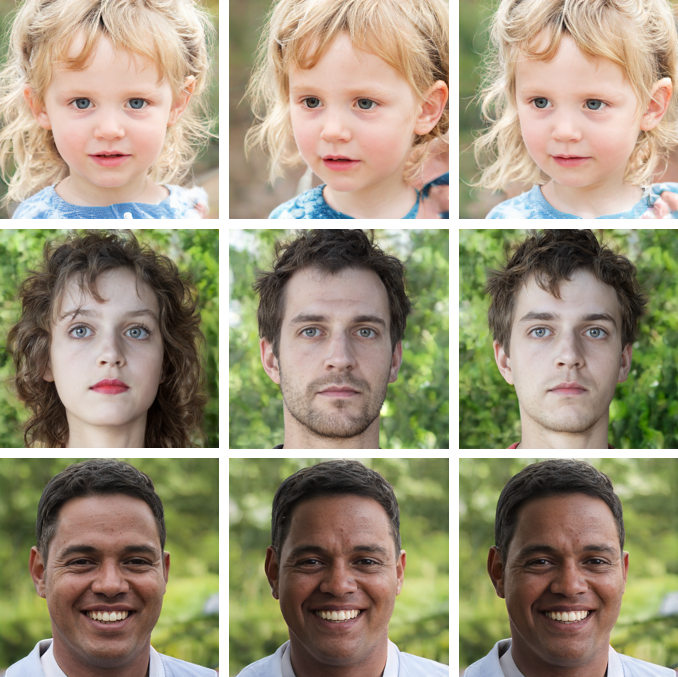}
        \caption{
        Effect of training conditional CNFs with single attribute versus simultaneously along multiple attributes. In this example, from top-to-bottom, the target edits: pose change only, gender change only, illumination change only. In the  single-attribute case~(second column), the edits result in changes along other attributes: additionally identity changes, additionally age changes, and additionally pose changes, respectively. In contrast, in the multi-attribute case~(right column), other attributes are better preserved.  }
        \label{fig:SGy1}
    \end{figure}
    
\begin{figure}[t!]
        \centering
        \includegraphics[width=\linewidth]{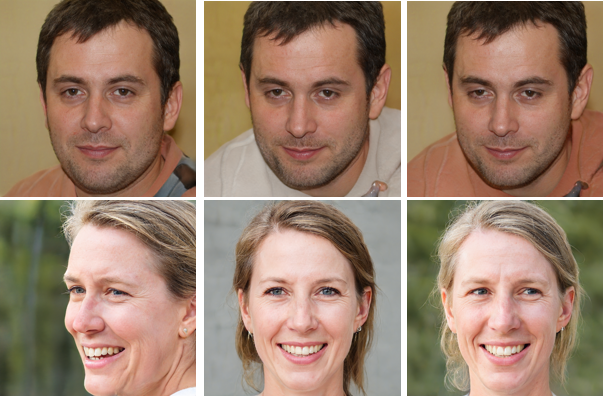}
        \caption{Importance of edit-specific subset selection. (Left)~Input image; (middle)~changes performed without edit-specific subset selection block; (right)~changes performed with edit-specific subset selection block. As seen, the subset selection results in better preservation of the background. Please refer to the text for details. }
        \label{fig:editspecific}
    \end{figure}
    
\begin{figure*}[t]
        \centering
        \includegraphics[width=0.90\linewidth]{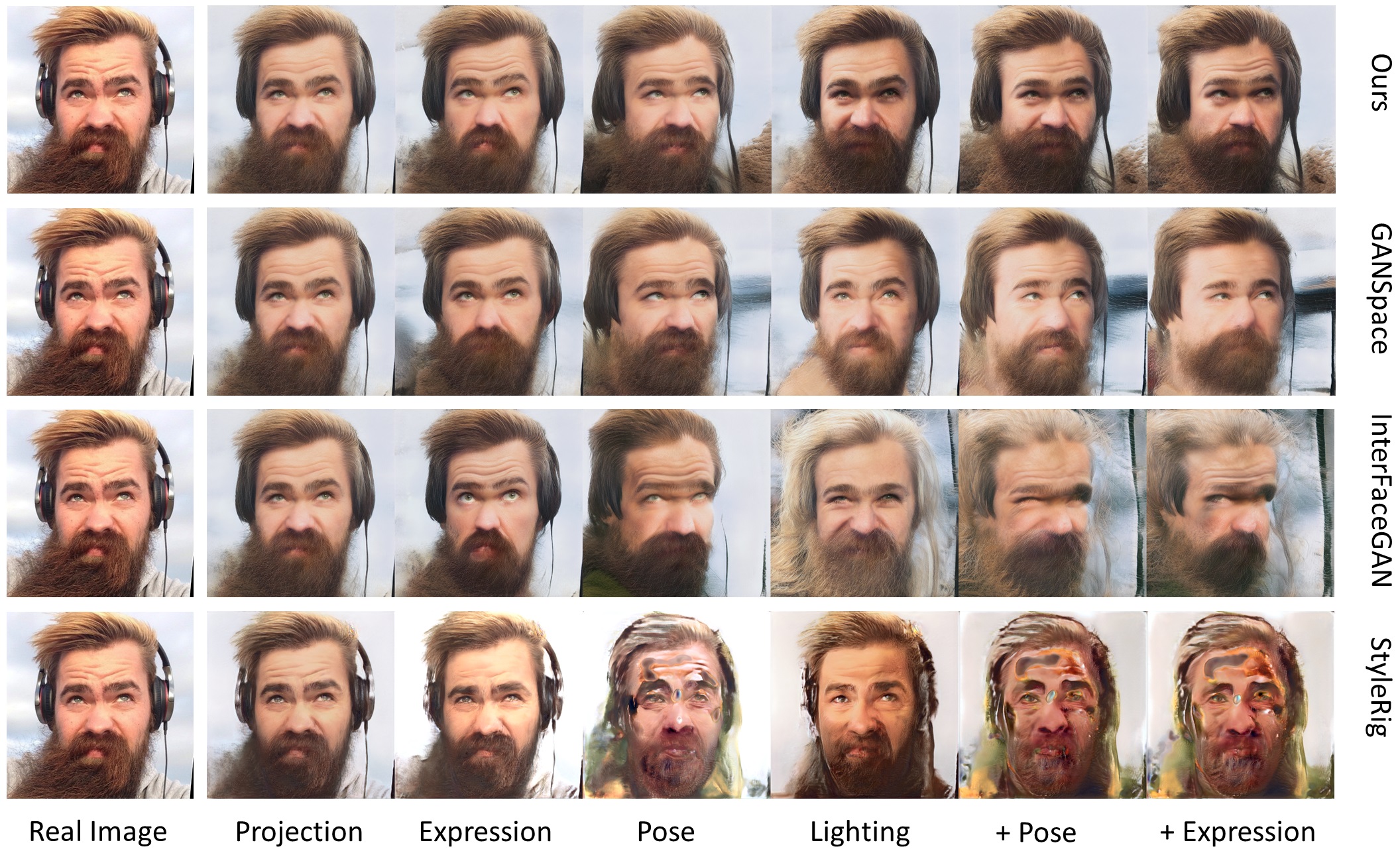}
        \includegraphics[width=0.90\linewidth]{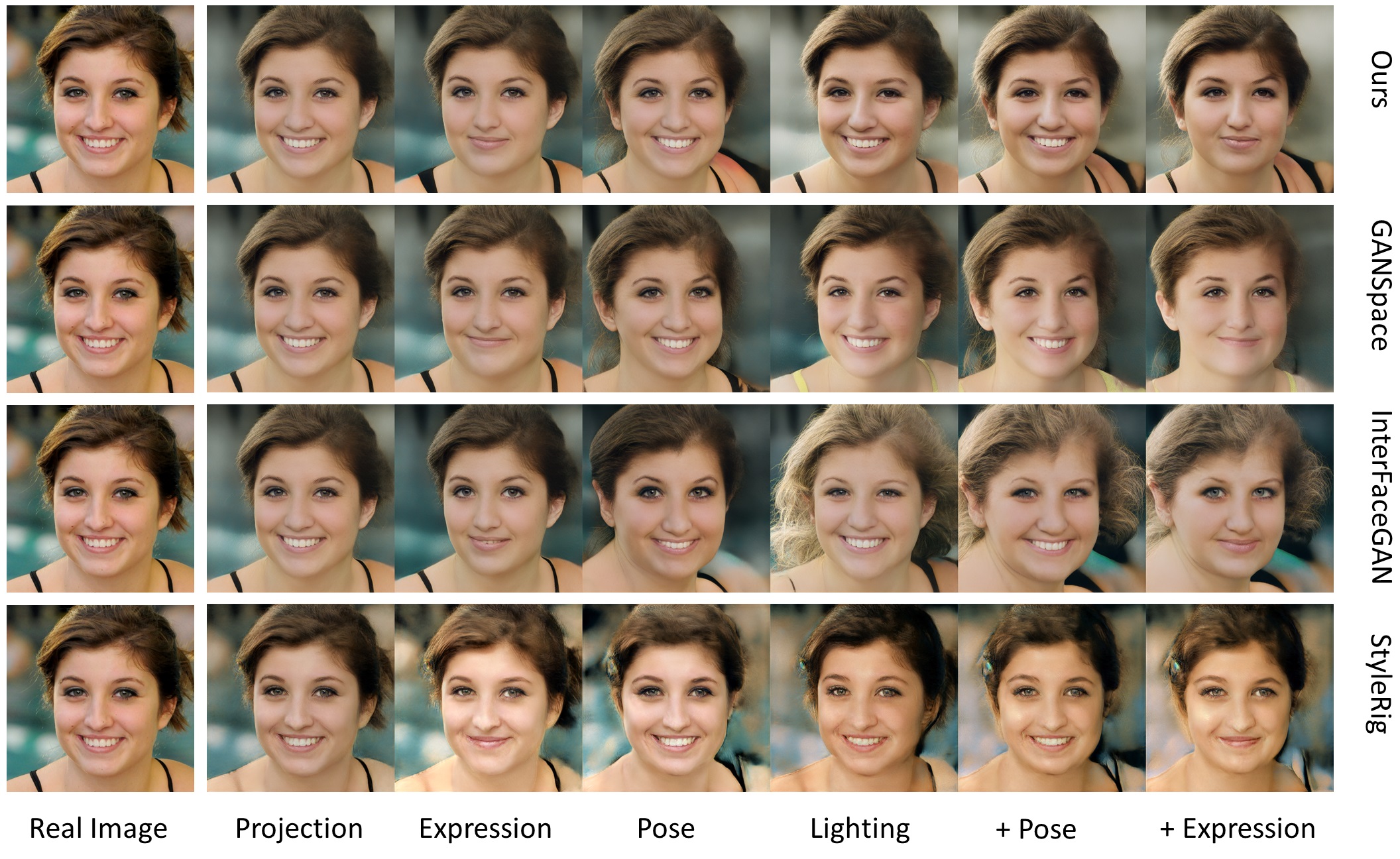}
        \caption{Real image editing comparison with competing methods. }
             \label{fig:realx}
    \end{figure*}
\begin{figure*}[ht]
        \centering
        \includegraphics[width=0.90\linewidth]{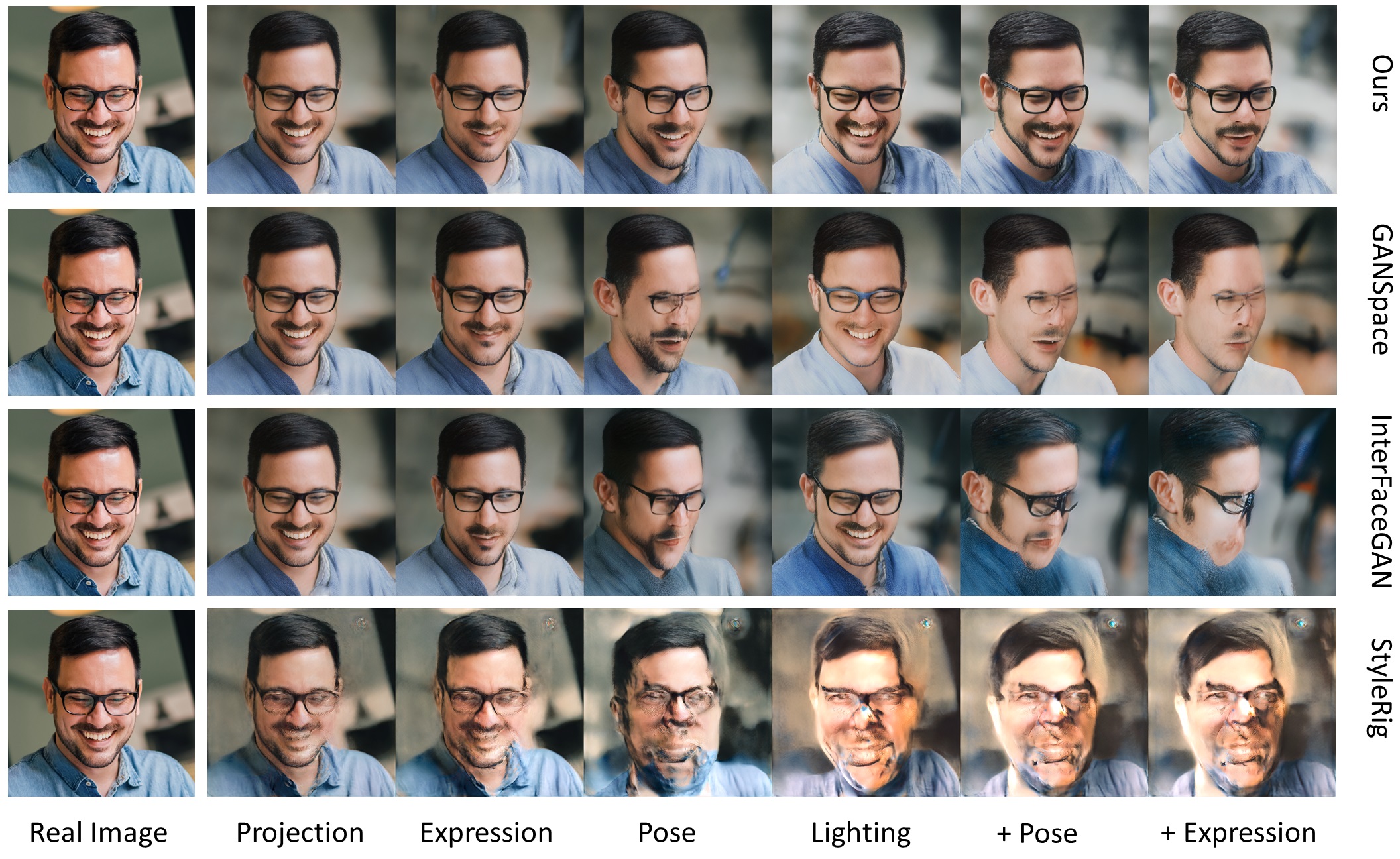}
        \includegraphics[width=0.90\linewidth]{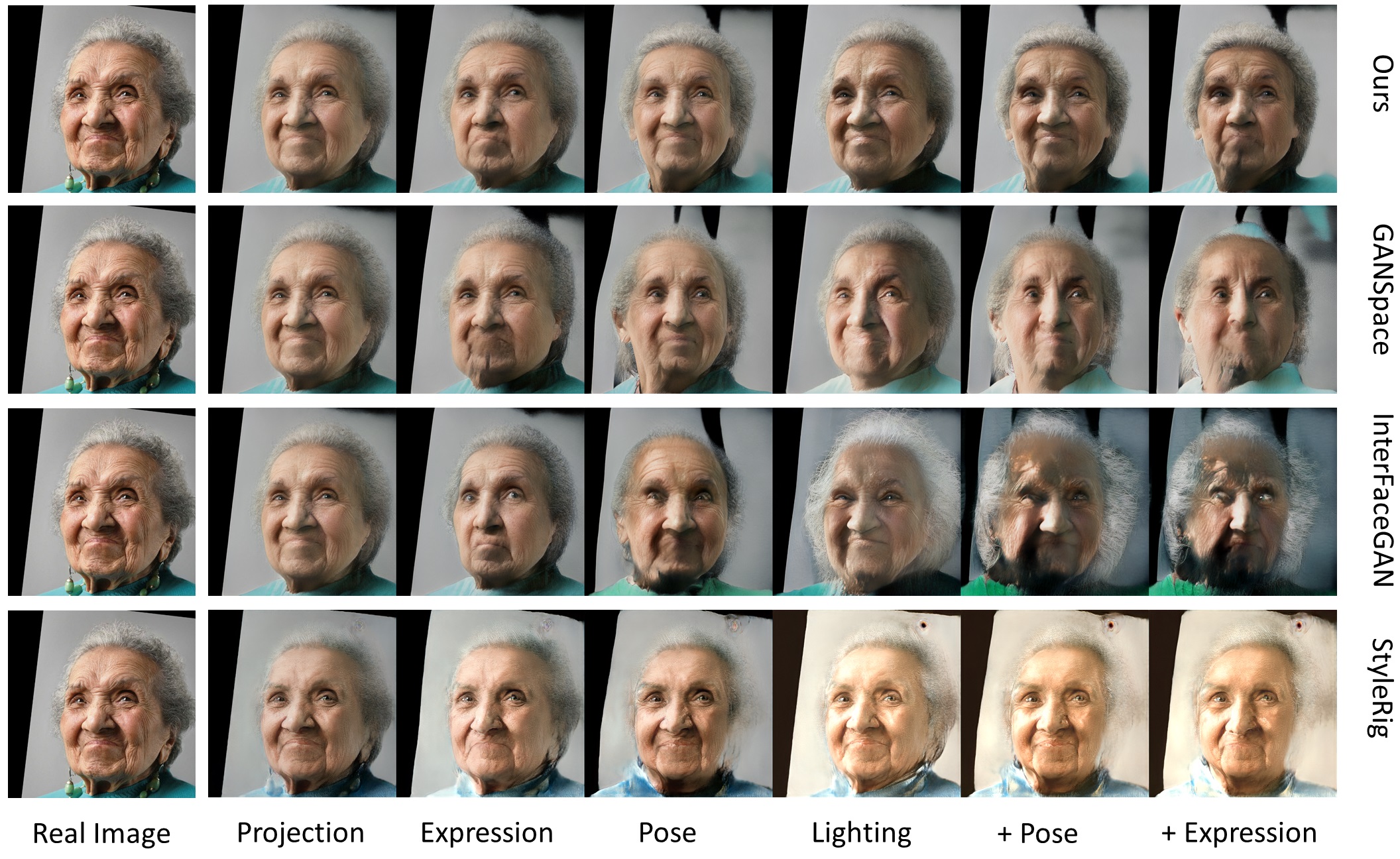}
        \caption{Real image editing comparison with competing methods (cont.).  }
        \label{fig:real}
    \end{figure*}

To assess the \textit{non-linearity} of the edit path, we compare the interpolation in the attribute domain ($a_t$) to the interpolation in the latent domain ($w$),  i.e., we linearly change the variable of the attribute that is fed to the flow model versus linear interpolation of the vector $w$ to $w'$. We sample 20 points along the interpolation paths of both scenarios and then compare the latents produced by both the methods. We compute mean of the norm of these difference vectors along the path. Sampling multiple edits produced by StyleFlow, we conclude that, on average, the linear interpolation in the $w$ domain differs from the attribute domain $a_t$ by a factor of $1.5$. In \revision{Figure~\ref{fig:interpol}, we compare the results of the non-linear path edits with the linear interpolation visually. Note here the final w' is obtained by subjecting StyleFlow to extreme edits. We notice that the non-linear path is able to retain hair style, clothes and head coverings for a larger extent along the path validating the improvement in the disentanglement.}

We perform sequential edits to the generated images as shown in \revision{supplementary materials Figure 1} i.e., `Pose', `Light' and `Expression'. In Table~\ref{tab:three}, we show the FID score for our method is relatively low (lower the better) than other methods.

Identity preservation results are presented in Table~\ref{tab:one}.
 The metrics show that our method outperforms others across all metrics and all edits. One exception are expression edits for which GANSpace also performs well. Also we evaluate the scores when all the edits are applied sequentially. Here, our method also shows superiority in quantitative evaluation. Moreover, we also compute the accuracy based on the final decision of the classifier of the two embeddings being the same face.

In Table~\ref{tab:two}, we show the cyclic edit consistency evaluation of our method and compare it with other methods. 
As shown in the Table~\ref{tab:two}, our editing method remains consistent under different permutations. We used mean (absolute) error across the respective attributes. \revision{Note that for GANSpace since only four disentangled edits match our attributes, i.e.,  gender, expression, pose and lighting, we are restricted in the comparison. In case of the expression, we notice that the attribute classifier outputs binary values, so the consistency scores for every method is zero. We make the same observation for gender change.}

\begin{table}%
\caption{Using FID~(Fr$\acute{e}$chet Inception Distance) score to compare realism of sequential edits using different methods. }
\label{tab:three}
\begin{minipage}{\columnwidth}
\begin{center}
\begin{tabular}{rlll}
  \toprule
  Sequential edit  & $FID \Downarrow$\\ \midrule
Image2StyleGAN   & 82.49   \\

 InterfaceGAN    & 67.08 \\
  GANSpace & 64.69 \\
  Our(V2)  & \textbf{53.15} \\
  \bottomrule
\end{tabular} 
\end{center}
\end{minipage}
\end{table}%

\begin{table}%
\caption{Identity preservation achieved by different methods as evaluated by a SOTA face classifier~\cite{Fr}. Please refer to the text for details. }
\label{tab:one}
\begin{minipage}{\columnwidth}
\begin{center}
\begin{tabular}{rrlllll}
  \toprule
  Edit  & Metric & $I2S$ & $IG$ & $GS$ &Ours(V1) & Ours(V2)\\ \midrule
  Light     & $C_s \Uparrow$ &0.910 &0.945 &0.942 &\textbf{0.958} &\textbf{0.963}\\
            & $E_d \Downarrow$&0.633 &0.508 &0.524 &\textbf{0.438} &\textbf{0.394}\\
  Pose     & $C_s \Uparrow$ &0.877 &0.940 &0.939 &\textbf{0.952} &\textbf{0.966}\\
            & $E_d \Downarrow$&0.748 &0.532 &0.526 &\textbf{0.466} &\textbf{0.400}\\
 Expression & $C_s \Uparrow$ &0.941 &0.946 &\textbf{0.973} &0.951 &\textbf{0.967}\\
            & $E_d \Downarrow$&0.534 &0.509 &\textbf{0.359} &0.472 &\textbf{0.388}\\
  All     & $C_s \Uparrow$ &0.870 &0.895 &0.902 &\textbf{0.923} &\textbf{0.941}\\
            & $E_d \Downarrow$&0.774 &0.690 &0.681 &\textbf{0.564} &\textbf{0.529}\\
            & $Acc \Uparrow$ &0.000 &0.300 &0.550 &\textbf{0.900} &\textbf{0.950}\\

  \bottomrule
\end{tabular}
\end{center}
\bigskip\centering
\footnotesize\emph{Notations: $I2S$ - Image2StyleGAN ; $IG$ - InterfaceGAN ; $GS$ - GANSpace ; \\  $C_s$ - Cosine Similarity ; $E_d$ - Euclidean Distance ; $Acc$: Accuracy. }

\end{minipage}
\end{table}%

\begin{table}[t]
\caption{Edit Consistency Evaluation (mean absolute error) to compare different methods under permutation of edit operations. }
\label{tab:two}
\begin{minipage}{\columnwidth}
\begin{center}
\begin{tabular}{rllll}
  \toprule
  Edit  & $I2S$ & $IG$ & $GS$ & Ours(V2)\\ \midrule
  Pose $(ep - pl)$    &4.68  &10.18 & 10.53 &\textbf{1.64}\\

  Light $(le - pl)$  &0.83 &0.66 & 0.58 &\textbf{0.53}\\
  
   Facial hair $(fl - pf)$    &0.31 &0.23 & -    &\textbf{0.19}\\

  \bottomrule
\end{tabular}
\end{center}
\bigskip\centering
\footnotesize\emph{Notations: $I2S$ - Image2StyleGAN ; $IG$ - InterfaceGAN ; $GS$ - GANSpace ; \\ e - expression ; p - pose ; l - light ; f - facial hair  }

\end{minipage}
\end{table}%

\begin{figure*}[t!]
        \centering
        \includegraphics[width=.8\linewidth]{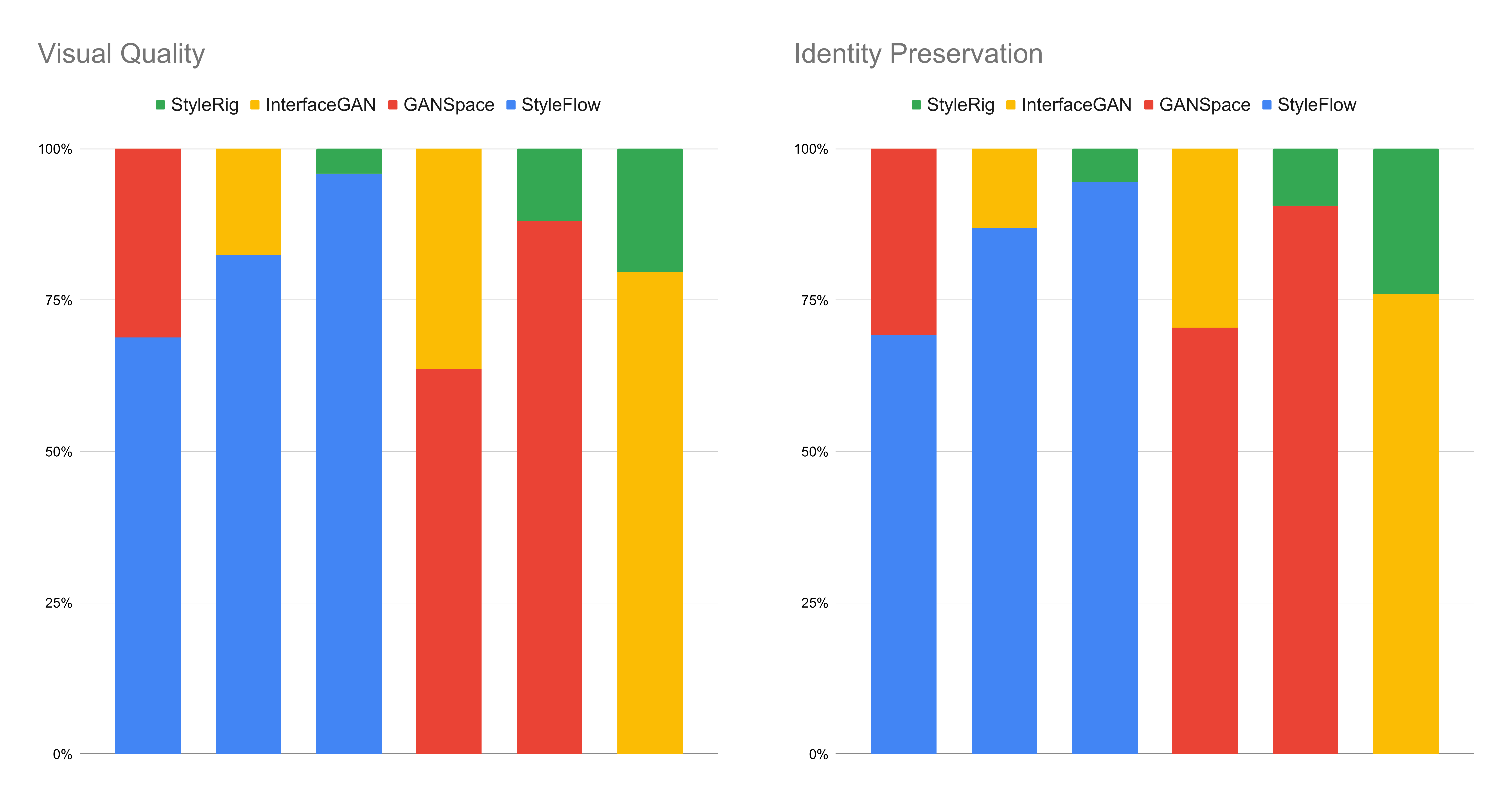}
        \caption{User study results assessing visual quality of the edits (left) and identity preservation (right). }
        \label{fig:vis1}
    \end{figure*}

\begin{table}%
\caption{Ablation Study of StyleFlow CNF functions.}
\label{tab:abb}
\begin{minipage}{\columnwidth}
\begin{center}
\begin{tabular}{rrllll}
  \toprule
  Stack & Parm & Infer & LL & Result\\ \midrule
  2 &565249 &0.40 &-4323537 & Overfit\\
  3 &846849 &0.37 &-4325796 & Overfit\\
 4 &1128449 &0.61 &-4327872 & Selected model\\
 6 &1691649 &0.21 &-4328470 & Faster alternative\\
  \bottomrule
\end{tabular}
\end{center}
\bigskip\centering
\footnotesize\emph{\revision{Notations: Stack - Number of stacked functions ; Parm - Number of parameters; \\  Infer - Inference time (sec); LL - Final Log likelihood (higher the better).} }

\end{minipage}
\end{table}%
\subsection{Choice of encoding and subset selection}

We evaluate the two design choices of joint attribute encoding and edit specific subset selection.  As explained, we perform joint attribute encoding to ensure the face identity is preserved during the Conditional Forward Editing (CFE). Figure~\ref{fig:SGy1} shows the variation of the proposed approach when attributes are trained jointly versus separately. The results show that in case of the joint encoding of the attributes, the identity of the face and the unedited attributes like hair style, age, and background are better preserved. 
We evaluate the effectiveness of the edit specific subset selection block in Figure~\ref{fig:editspecific}. We notice that the edits done with the V2 framework performs high quality edits producing images with comparable the skin tone, background, and clothes with respect to the source image. 
\revision{We also provide an ablation study evaluating the architecture with regards to different numbers of CNF function blocks in Table~\ref{tab:abb}. }
    
\subsection{User Study}

\revision{In order to evaluate the visual quality and the identity preservation of the images after the performed edits, we setup a user study. In a pairwise test setup, we asked, `Which of the two edited images better preserves the identity of the person in the original image ?' and `Choose which edited image among the two is more realistic and of higher visual quality?' We compare with 4 different methods, i.e., InterfaceGAN, GANSpace, StyleRig and Styleflow. We consider 5 type of edits common to all the methods expression, pose, lighting, and sequential Edits in the form of lighting + pose and lighting + pose + expression. 
We collected 18 diverse real images for the evaluation (4 different examples shown in  Figures~\ref{fig:realx} and \ref{fig:real}).
Since we perform pairwise comparisons (one edit at a time), our evaluation contains 1080 different comparisons per task. We divide these comparisons into 18 comparisons of 60 image pairs per task. We asked 25 people to perform both the tasks. Figure~\ref{fig:vis1} shows the results of the user study comparing 2 methods at a time and aggregating all the edit scores. The results show that StyleFlow  outperforms others in terms of visual quality and identity preservation. Table~\ref{tab:real} shows quantitative results of identity preservation on the real dataset. While StyleRig has a better evaluation of the initial projected image, after editing our method is better on all edit types, except for expression edits. We also encourage the reader to visually evaluate the 18 example edits included in supplemental.}

\begin{table}%
\caption{Identity preservation achieved by different methods as evaluated by a SOTA face classifier~\cite{Fr} on real image dataset. }
\label{tab:real}
\begin{minipage}{\columnwidth}
\begin{center}
\begin{tabular}{rrllll}
  \toprule
  Edit  & Metric & $IG$ & $GS$ & $SR$ &Ours \\ \midrule
  Projected Image  & $C_s \Uparrow$ &0.980 &0.980 &\textbf{0.988} &0.980 \\
            & $E_d \Downarrow$&0.283 &0.283 &\textbf{0.210} &0.283 \\
 
 Expression & $C_s \Uparrow$ &0.963 &\textbf{0.969} &0.968 &0.964 \\
            & $E_d \Downarrow$&0.387 &0.356 &\textbf{0.353} &0.380 \\

Pose     & $C_s \Uparrow$ &0.899 &0.961 &0.954 &\textbf{0.967} \\
            & $E_d \Downarrow$&0.468 &0.390 &0.414 &\textbf{0.364} \\
            
 Light     & $C_s \Uparrow$ &0.952 &0.902 &0.955 &\textbf{0.962} \\
            & $E_d \Downarrow$&0.440 &0.458 &0.414 &\textbf{0.394} \\
            
 Light + Pose   & $C_s \Uparrow$ &0.874 &0.893 &0.941 &\textbf{0.951} \\
              & $E_d \Downarrow$&0.569 &0.500 &0.473 &\textbf{0.444} \\  
              
 Light + Pose & $C_s \Uparrow$ &0.765 &0.884 &0.925 &\textbf{0.941}\\
 + Expression & $E_d \Downarrow$&0.641 &0.533 &0.540 &\textbf{0.484}\\

  \bottomrule
\end{tabular}
\end{center}
\bigskip\centering
\footnotesize\emph{Notations: $IG$ - InterfaceGAN ; $GS$ - GANSpace ;  $SR$ - StyleRig ; \\  $C_s$ - Cosine Similarity ; $E_d$ - Euclidean Distance. }

\end{minipage}
\end{table}%

\subsection{Qualitative editing results on cars}
\begin{figure}[b!]
        \centering
        \includegraphics[width=\linewidth]{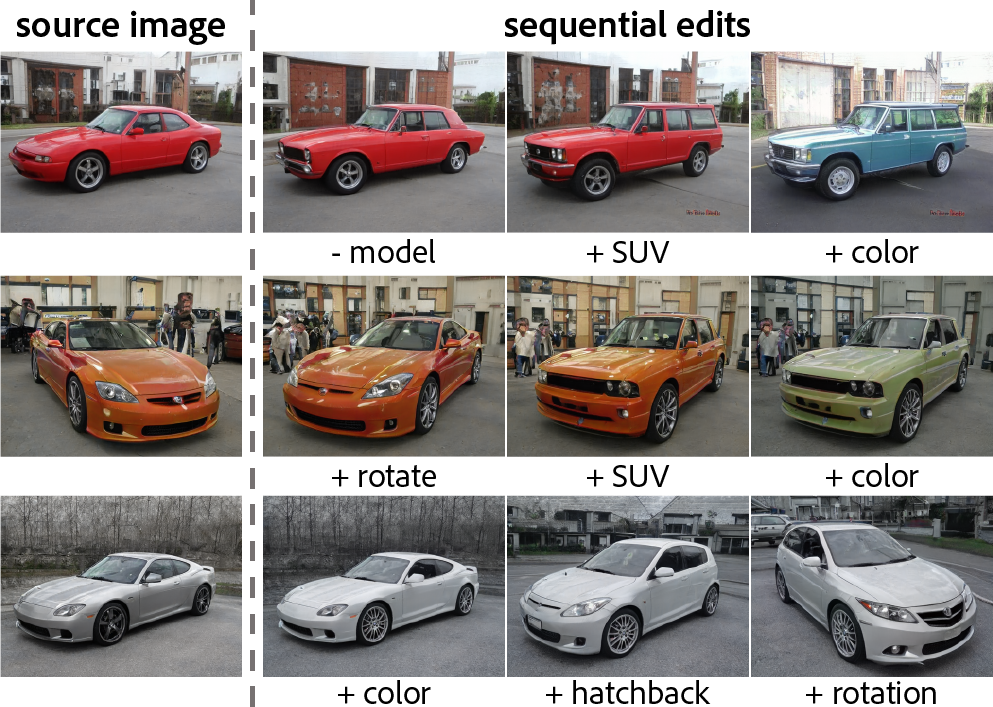}
        \caption{Sequential attribute-conditioned edits using StyleFlow on the car latent space of StyleGAN.}
        \label{fig:car2}
    \end{figure}
We also show the editing results of our framework with StyleGAN2 trained on the LSUN-Car dataset, Figure~\ref{fig:car2} shows qualitative results of our framework. We show sequential edits including SUV/Hatchback conversion, rotation, and color change. We use a fine-tuned ResNet-152~\cite{he2016deep} model trained on Stanford Cars~\cite{KrauseStarkDengFei-Fei_3DRR2013} to create the attributes. \revision{For car manipulation, we used a car recognition model~\cite{spec}  with 95\% classification accuracy, we report the accuracy for the Hatchback/SUV conversion as 80\% and the color as 100\%. For the rotation, there is no precise model in the literature to evaluate the scores quantitatively, hence we only show multiple visual examples in the supplementary video.}  

\subsection{Attribute-conditioned sampling}

We show the results of the conditional sampling of StyleGAN2 in Figure~\ref{fig:sampling}.  In the first row, for instance, we sample females of different age groups with glasses and fixed pose. Note that during the sampling operation we resample $z$ to infer vectors in $w$ and keep a set of attributes fixed. Apart from the quality of the samples we find the diversity of the samples to be also high.

\subsection{StyleFlow editing interface}
\begin{figure}[h!]
        \centering
        \includegraphics[width=\columnwidth]{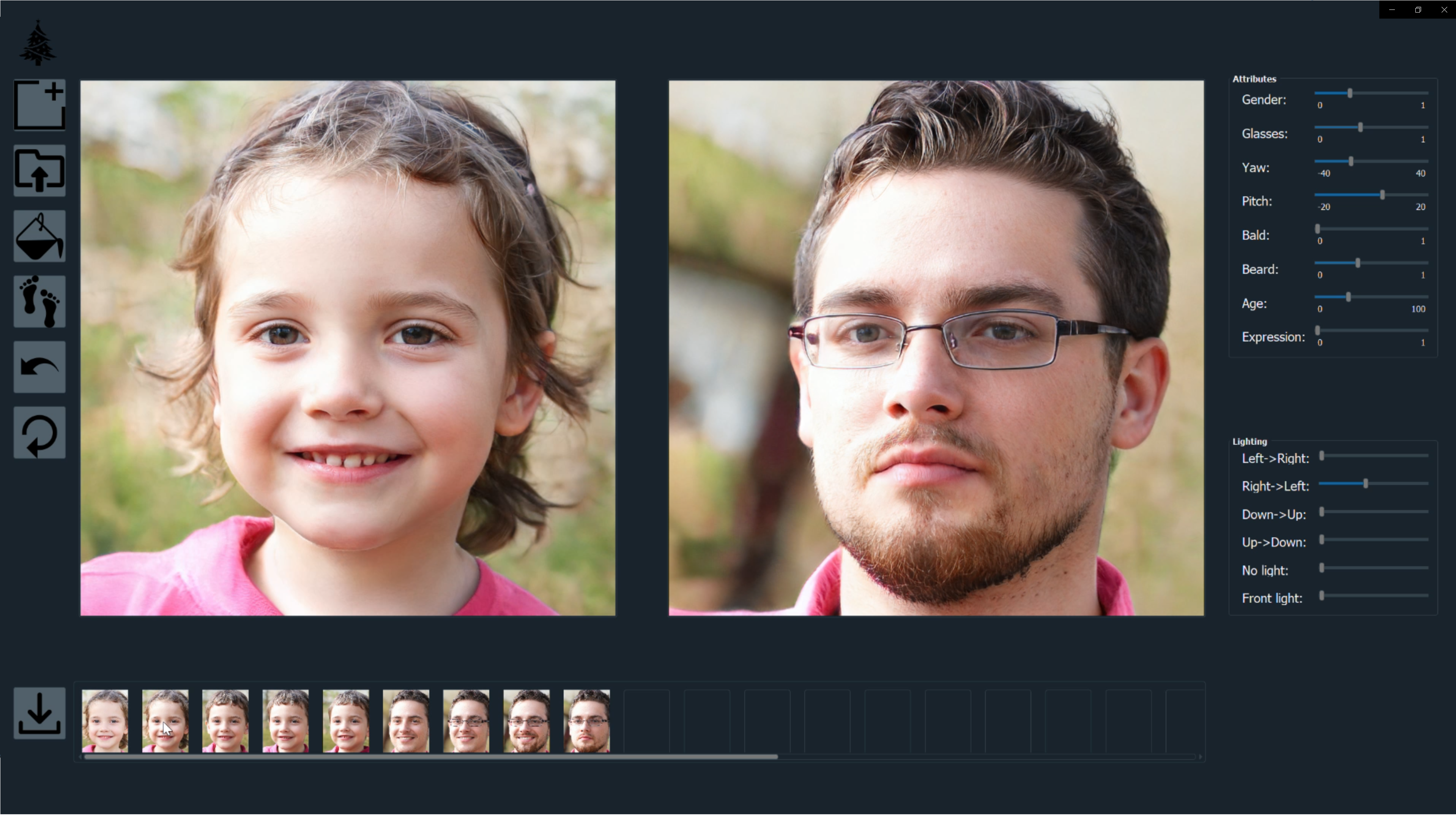}
        \caption{StyleFlow user interface. Please refer to the supplementary video. }
        \label{fig:ui}
    \end{figure}

To enable interactive editing, we developed an image editing interface (see Figure~\ref{fig:ui}) that allows a user to select a given real or generated image and perform various edits with the help of interactive sliders. For sequential edits, the checkpoint images are saved in a panel so that a user can revisit the changes made during the interactive session. 


\section{Discussion}


Here we describe different aspects of the different editing methods and how the design choices effect the results.

\subsection{Edit specific subset selection}
GANSpace and StyleFlow both use edit specific subset selection. This is an advantage over InterfaceGAN (and possibly StyleRig) and contributes to disentanglement in our results. This might be the reason why GANSpace seems to perform better than InterfaceGAN.

\subsection{Conditioning editing direction on the starting image}
One key difference between our method and others such as Image2StyleGAN, InterfaceGAN, and GANSpace, is that the editing direction depends on the starting latent. These other three competing methods compute a single editing direction $\mathbf{d}$ for an attribute (e.g., expression or pose) and apply this same editing direction to all starting latents $\mathbf{w}$ or $\mathbf{w+}$ by scaling and adding the vector $\mathbf{d}$.
In contrast, our methods as well as StyleRig compute an edit direction that depends on the starting latent. Our results show that this design choice contributes to better disentanglement.

\subsection{Supervised vs. unsupervised edit discovery}

Our method as well as StyleRig, InterfaceGAN, and Image2StyleGAN are supervised. They need to have a training corpus of latent vectors / images that are labeled with attributes. In contrast,  GANSpace discovers edits in an unsupervised manner. GANSpace finds a large number of edits that could be interesting and then labels them semantically using visual inspection. Our results show that the unsupervised discovery of edit directions works reasonably well. It sometimes leads to edits that are not identity preserving, as multiple changes are entangled together. However, GANSpace has the ability to discover cool edits for which labels might not be available. One example is the expression where the mouth forms an O-shape.

\subsection{Binary attributes vs. continuous attributes}

Our method can create edits by specifying a continuous parameter for a given attribute. InterfaceGAN and GANSpace just give an edit direction and the user has to manually tune the scaling parameter to control the strength of an edit. That makes the edits not directly comparable and we have to manually tune this scaling parameter of these other methods to make it approximately match the strength of our edit.

\subsection{Linear vs. non-linear interpolation path}

The edit paths in latent space of ours and StyleRig are non-linear. This is in contrast to InterfaceGAN and GANSpace that use linear trajectories. Our evaluation indicates that this difference is not the only factor in explaining the edit quality of our method. It is also possible to setup our method approximating the nonlinear edit path by a linear one to achieve better results than all competing works.

\subsection{Training one attribute at a time vs. multiple attributes}

In our results we show that the quality of edits improves if we train StyleFlow using all attributes at the same time. This helps disentanglement and provides higher quality edits than training a network for each attribute separately.
This is in contrast to StyleRig. In the StyleRig paper the results are shown by training for each edit separately. In additional materials, the authors demonstrate that it is also possible to train for all edits jointly with some loss in quality. We believe that our behavior is more intuitive.

\subsection{Individual vs. sequential edits}

Previous work often focuses on applying a single edit to a starting image. In our experience, the quality of an editing framework becomes more obvious when applying sequential edits. Small errors in disentanglement accumulate and the initial face much easier loses its identity when subsequent edits are applied. While we also show results for individual edits, we primarily focus on sequential edits in the paper to provide a more challenging evaluation setup.

\subsection{What type of edits are possible?}

The quality of edits depends on the availability of good attribute labels and a good training dataset for StyleGAN. In general, the face dataset has the highest quality attributes and the highest quality latent space. Also, StyleRig can only work on faces. We therefore focus mainly on faces in our evaluation. However, there is nothing specific to faces and our work can be applied to other suitable dataset.
For faces, InterfaceGAN and StyleFlow can produce edits for the same set of attributes. However, StyleRig does not include several common attributes such as hair length, gender, eyeglasses, age. 
We are therefore also restricted in our evaluation to a subset of the attributes when comparing to StyleRig.

\subsection{Editing real  vs. synthetic images}

In principle, there is no difference between editing real images and editing synthetic images. When editing real images, it becomes important to have a good projection method into latent space. We use a reimplementation of Image2StyleGAN for StyleGAN2. This method seems to be sufficient to get good edits, but multiple researchers are working on improved embedding algorithms. Analyzing the compound effect of embedding and editing is beyond the scope of this paper. We just would like to note that it is essential to embed into $\mathbf{w+}$ latent space and not into 
$\mathbf{w}$ latent space as for example proposed by the StyleGAN2 paper. All methods can work with real as well as synthetic images. However, StyleRig seems to have the most problems working with real images and all results shown in their paper are on synthetic images only. 

\section{Conclusion}
We presented StyleFlow, a simple yet robust solution to the conditional exploration of the StyleGAN latent space. We investigated two important subproblems of attribute-conditioned sampling and attribute-controlled editing on StyleGAN using conditional continuous normalizing flows. As a result, we are able to sample high quality images from the latent space given a set of attributes. Also, we demonstrate fine-grained disentangled edits along various attributes, e.g., camera pose, illumination variation, expression, skin tone, gender, and age for faces. \revision{The real face editing of our framework is demonstrated to have unmatched quality than the concurrent works.} The qualitative and quantitative results show the superiority of the StyleFlow framework over other competing methods. 

We identified three major limitations of our work. First, our work relies on the availability of attributes. These attributes might be difficult to obtain for new datasets and could require a manual labeling effort.
Second, great results are only achievable with StyleGAN trained on high quality datasets, mainly FFHQ. It would be good to have different types of datasets in similar quality, e.g., buildings or indoor scenes, to better evaluate our method. The lack of availability of very high quality data is still a major limitation for evaluating GAN research.
Third, \revision{the real image editing sometimes produce some artifacts compared to the synthesized images.} While the quality of these edits in our framework are still better than competing work, a better understanding of this problem \revision{ and better projection algorithms require} further research.
We suggest this line of investigation as the most rewarding avenue of future work. In addition, it would be interesting to develop extensions to other attributes. In this context, it would be interesting to analyze what attributes are even captured by a GAN model. Maybe a combination of GANSpace to discover attributes and our method to encode conditional edits could be developed in the future.
\begin{figure*}[h!]
        \centering
        \includegraphics[width=\linewidth]{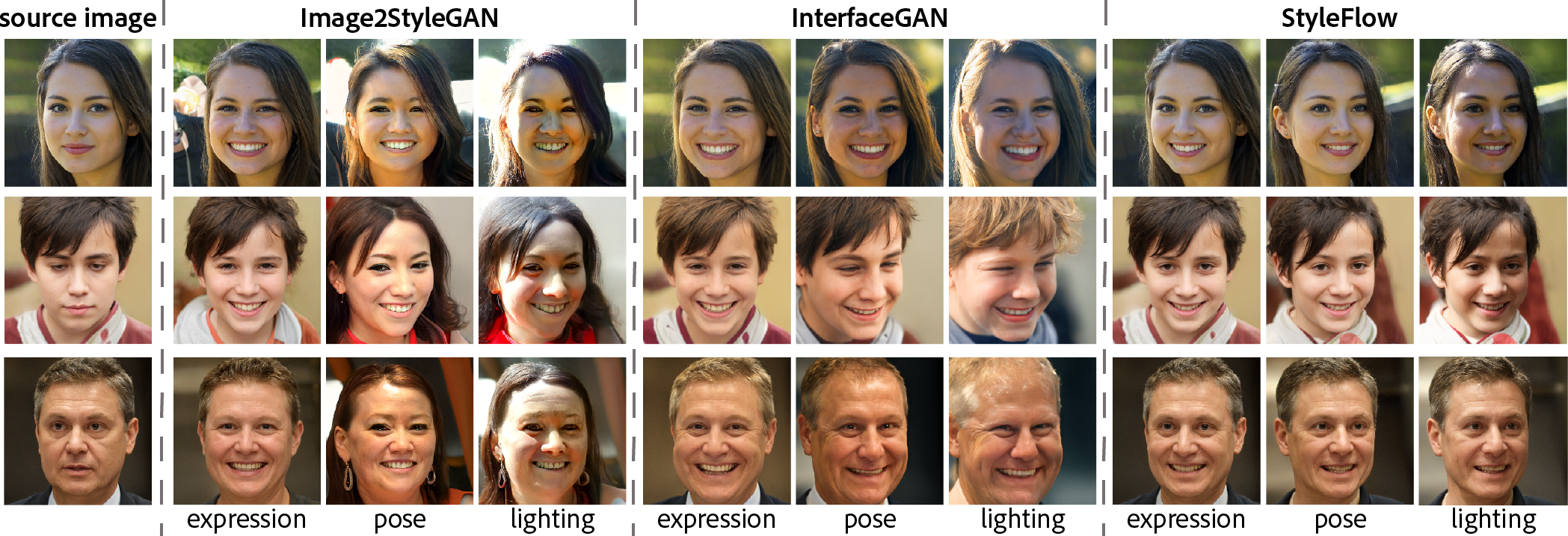}
        \caption{Comparison of StyleFlow with other contemporary systems 
        Image2StyleGAN~\cite{abdal2019image2stylegan} and  InterfaceGAN~\cite{shen2019interpreting}.}
        \label{fig:compare}
    \end{figure*}

\section{Appendix}
\subsection{Qualitative comparisons on generated images}
Figure~\ref{fig:compare} compares the quality of the edits. Here, we subject the methods of Image2StyleGAN\cite{abdal2019image2stylegan}, InterfaceGAN~\cite{shen2019interpreting}, and our StyleFlow to extreme attribute conditions and perform sequential edits on the images. We consider three primary edits of pose, expression, and lighting. The figure shows that while Image2StyleGAN suffers and drives the image out of the distribution due to its usage of $W+$ space for the edit computation, InterfaceGAN's conditional manipulation produces relatively better image samples. However, preserving face identity still remains a major issue. In contrast, \name handles the sequential edits producing high quality output and preserves facial features.

\begin{figure*}[t]
        \centering
        \includegraphics[width=\linewidth]{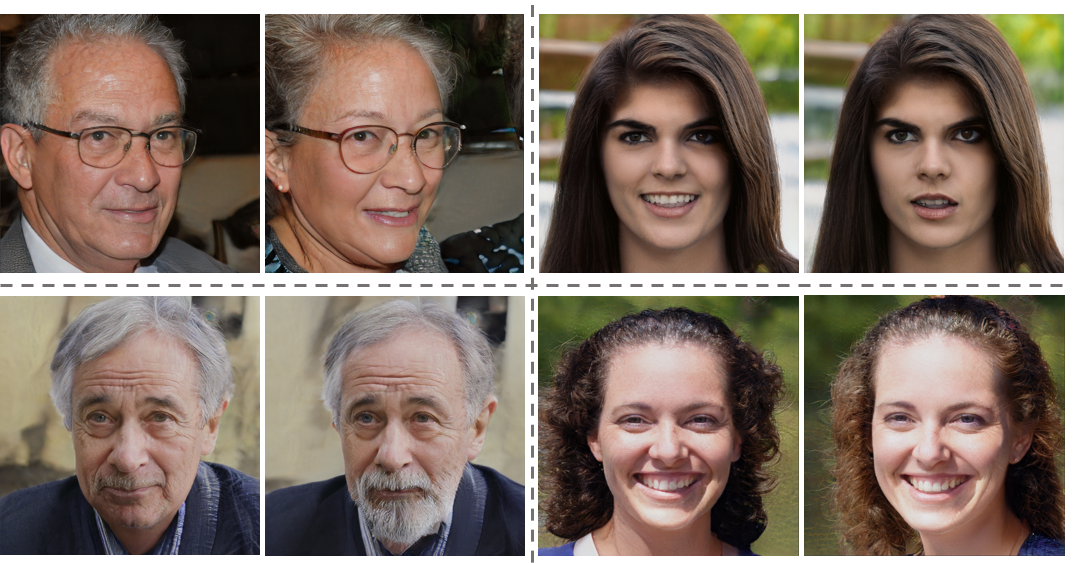}
        \caption{
        Attribute-conditioned edits on images sampled from the StyleGAN1 latent space. The attributes are gender~(top-left), expression~(top-right), facial hair~(bottom-left), and 
        pose~(bottom-right). }
        \label{fig:SG1}
    \end{figure*}
    
\subsection{Compatibility with StyleGAN1}

To demonstrate the compatibility of our work with the older StyleGAN1, we show the results of selected edits in Figure~\ref{fig:SG1}. Despite the more entangled latent space, our method is able to perform well. 

 \begin{figure*}[t]
        \centering
        \includegraphics[width=\linewidth]{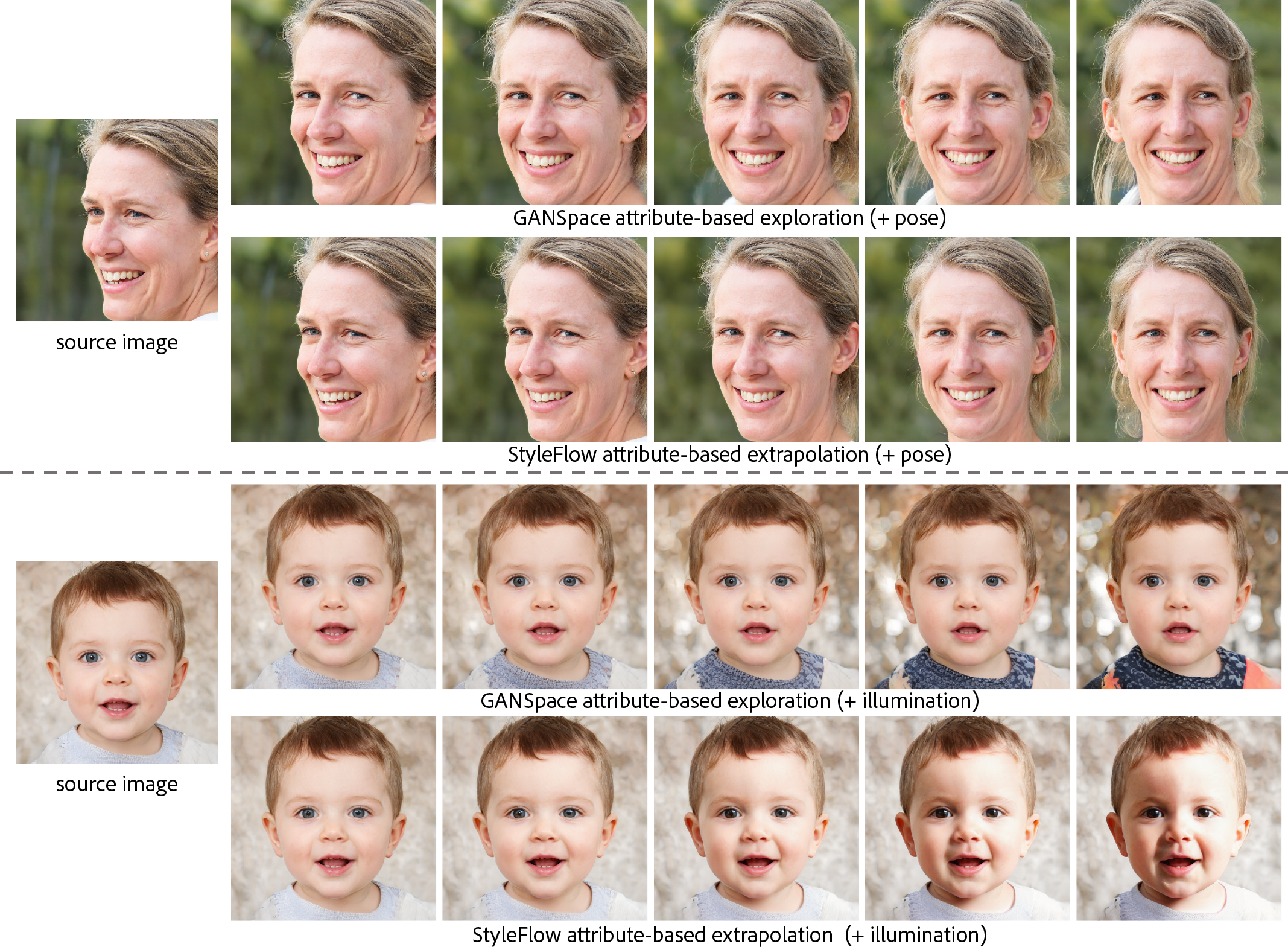}
        \caption{Comparison of StyleFlow with GANSpace~\cite{harkonen2020ganspace}. 
        The upper example shows change of pose, while the lower example shows change of illumination. Compare the quality of disentanglement achieved by the two methods. Please note that GANSpace, being an unsupervised method, produces a fixed set of (PCA) axis that may not correspond to target semantic directions. Finally, our conditional CNF being nonlinear, StyleFlow can expose non-linear paths in the latent space that better satisfies attribute changes compared to those using linear PCA axes produced using GANSpace.  
        }
        \label{fig:ganspace}
    \end{figure*}

\subsection{StyleFlow vs GANSpace}
Figure~\ref{fig:ganspace} shows a visual comparison of our method with the GANSpace method.  Here we compare the transition results produced by GANSpace. Notice, in the top sequence of the figure, the transition fails and drastically changes the gender from female to male, while our results are gender preserving. Moreover, we notice that the edits computed by GANSpace do not work in all scenarios. In the lower sequence of Figure~\ref{fig:ganspace}, we show a failure case of lighting edit. We attribute these failure cases to the fact that the GANSpace edits, although very interesting, are still linear in nature and do not depend on the current identity of the face. Here the results are shown till 2$\sigma$. Note that GANSpace, being unsupervised, cannot control which attributes or combination of attributes are discovered as PCA axes. In contrast, in StyleFlow we  directly learn nonlinear mapping between GAN latent space and targeted attribute variations. Nevertheless, GANSpace does not need image annotations which is an advantage of the method when working with new datasets.


\bibliographystyle{ACM-Reference-Format}
\bibliography{references}
\end{document}